%% file: main.tex
\documentclass{article} 
\usepackage[preprint]{neurips_2025}

\input{math_commands.tex}

\usepackage[T1]{fontenc}
\usepackage{pifont}
\usepackage{bbding} 

\newcommand{\ProjectName}{Sci-PRM}

\title{\ProjectName: A Tool Aware Process Reward Model for Scientific Reasoning Verification}

\author{
\textbf{Xiangyu Zhao\textsuperscript{1,2}, 
Henry Hengyuan Zhao\textsuperscript{3}, 
Yiheng Wang\textsuperscript{2,4}, 
Wanghan Xu\textsuperscript{2,4}, 
Yuhao Zhou\textsuperscript{2,5},} \\
\textbf{Qinglong Cao\textsuperscript{2,4}, 
Zhiwang Zhou\textsuperscript{2,6}, 
Lei Bai\textsuperscript{2}, 
Wenlong Zhang\textsuperscript{2 \Envelope}, 
Xiao-Ming Wu\textsuperscript{1 \Envelope}}
\\
\textsuperscript{1}The Hong Kong Polytechnic University \quad 
\textsuperscript{2}Shanghai AI Lab \quad 
\textsuperscript{3}National University of Singapore \\ 
\textsuperscript{4}Shanghai Jiao Tong University \quad 
\textsuperscript{5}Sichuan University \quad 
\textsuperscript{6}Tongji University
\\
\texttt{xiang-yu.zhao@connect.polyu.hk, zhaohengyuan99@gmail.com} \\
\texttt{zhangwenlong@pjlab.org.cn, xiao-ming.wu@polyu.edu.hk}
}

\usepackage{pdflscape}

\usepackage{textcomp}
\usepackage{rotating}

\usepackage{setspace}
\usepackage{microtype} 

\usepackage{graphicx}
\usepackage{subcaption}
\usepackage{caption}

\usepackage{booktabs}
\usepackage[table]{xcolor}

\usepackage{array}
\usepackage{threeparttable}
\usepackage{multirow}

\usepackage{amsmath}
\usepackage{siunitx}

\usepackage{enumitem}
\usepackage{float}
\usepackage{seqsplit}
\usepackage{framed}

\usepackage{tikz}

\usepackage{ragged2e}
\usepackage[most]{tcolorbox}

\usepackage{makecell}
\usepackage{tabularray}    

\usepackage[symbol]{footmisc}
\newcolumntype{Y}{>{\RaggedRight\arraybackslash}X}

\usepackage{hyperref}
\hypersetup{
    colorlinks=true,
    linkcolor=blue, 
    citecolor=blue,  
    filecolor=black,
    urlcolor=blue    
}

\usepackage{tocloft}

\usepackage{etoolbox}
\usepackage{arydshln}

\definecolor{promptblue}{RGB}{183,226,252}

\newcommand\model{Sci-PRM}

\definecolor{toolbg}{RGB}{235, 242, 250} 
\definecolor{oursbg}{RGB}{230, 255, 230}

\definecolor{bg_gray}{RGB}{245,245,245}
\definecolor{code_bg}{RGB}{250,250,250}
\definecolor{success_green}{RGB}{25, 135, 84}
\definecolor{fail_red}{RGB}{220, 53, 69}

\newtcolorbox[auto counter, number within=section]{promptbox}[2][]{%
  colback=white,
  colframe=promptblue,        
  coltitle=black,
  fonttitle=\normalsize,
  width=\linewidth,              
  arc=1mm,
  boxrule=0.5mm,
  left=1.2mm,right=1.2mm,top=1mm,bottom=1mm, 
  boxsep=1mm,
  title={#2},
  breakable,
  enhanced,
  #1
}

\newcommand{\statusbadge}[1]{%
  \ifstrequal{#1}{Valid}{\textcolor{success_green}{\textbf{[Valid]}}}{%
  \ifstrequal{#1}{Invalid}{\textcolor{fail_red}{\textbf{[Invalid]}}}{%
  \textbf{[#1]}}}%
}

\makeatletter
\patchcmd{\@tocline}
    {\hfil}
    {\leaders\hbox{\hfil}\hfil}
    {}{}
\makeatother

\begin{document}
\sloppy 
\maketitle


\begin{figure}[h!]
\centering
\includegraphics[width=0.96\textwidth]{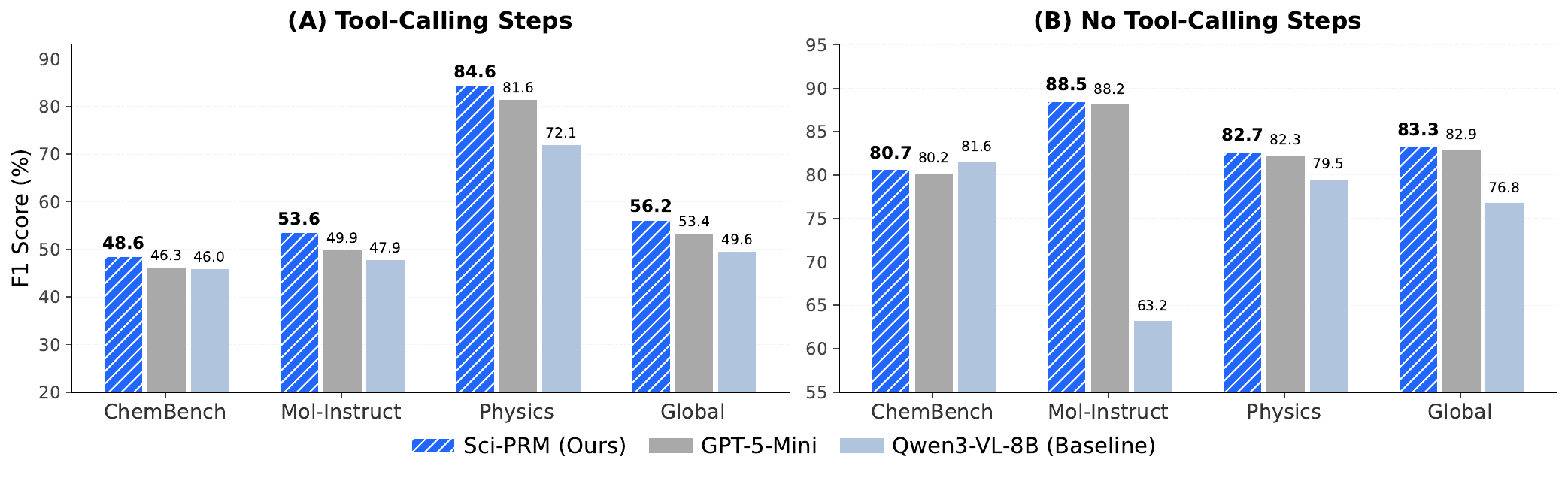}
\vskip -2mm
\caption{Step-level F1 scores of different judge models used as process rewards on four scientific benchmarks, separated by whether they contain tool-calling in current steps. The SoTA proprietary model GPT-5-Mini shows only minor gaps on non–tool-calling steps but degrades substantially on tool-calling steps. In contrast, \model{} (Ours), trained on our curated data, maintains high accuracy on both types of steps, highlighting its advantages and specialization in scientific domains.}
\label{fig:teaser}
\end{figure}



\input{sections/0_abs}
\input{sections/1_intro}
\input{sections/2_relatedwork}
\input{sections/3_method}

\input{sections/4_exp}
\input{sections/5_conclusion}



\bibliography{main}
\bibliographystyle{unsrt}

\clearpage
\input{sections/6_appendix}

\end{document}

%% file: math_commands.tex
\usepackage{amsmath,amsfonts,bm}

\def\eqref#1{equation~\ref{#1}}

\def\1{\bm{1}}

\DeclareMathAlphabet{\mathsfit}{\encodingdefault}{\sfdefault}{m}{sl}
\SetMathAlphabet{\mathsfit}{bold}{\encodingdefault}{\sfdefault}{bx}{n}



%% file: sections/0_abs.tex
\begin{abstract}
While Process Reward Models (PRMs) have achieved remarkable success in mathematical reasoning, their application in complex scientific domains—such as biology, chemistry, and physics remains largely unexplored. Scientific problems demand not only logical rigor but also factual consistency and the precise usage of domain-specific tools, areas where current models often suffer from hallucinations and lack of verification. In this paper, we first construct SCIPRM70K, a large-scale dataset featuring ``Chain-of-Tool'' trajectories that explicitly interleave reasoning with the execution of scientific tools. Building upon this, we train an efficient reward model called Sci-PRM to provide fine-grained supervision on tool selection, execution accuracy, and result interpretation at each step in one inference. Experiments demonstrate that Sci-PRM significantly enhances foundation models in two key aspects: (1) it enables effective test-time scaling via Best-of-N selection; and (2) when integrated into Reinforcement Learning, it serves as a dense reward signal that mitigates the critical issue of advantage disappearance, allowing the model to break through existing performance ceilings. \model{} is publicly available to foster further research and innovation: \url{https://github.com/InternScience/Sci-PRM}.
\end{abstract}

%% file: sections/1_intro.tex
\section{Introduction}

Large language models (LLMs) \citep{gpt-5, Gemini3, qwen3} are increasingly used as evaluators to provide training and test-time feedback, a paradigm often referred to as \emph{LLM-as-a-Judge}. Within this line of work, reward modeling approaches typically fall into two categories: \emph{outcome-supervised reward models} (ORMs), which assess only the final answer, and \emph{process-supervised reward models} (PRMs), which provide supervision over intermediate reasoning steps. ORMs are appealing for their simplicity and scalability, but they can assign high reward to outputs that are correct for the wrong reasons \citep{wen2025language, weng2024rewardhack}, an incidence of ``false positives'' \citep{yuan2025curingmiraclestepsllm} (e.g., reward hacking, miracle steps), thereby reinforcing unreliable responses. 


\begin{figure*}[t]
	\centering
	\includegraphics[width=\textwidth]{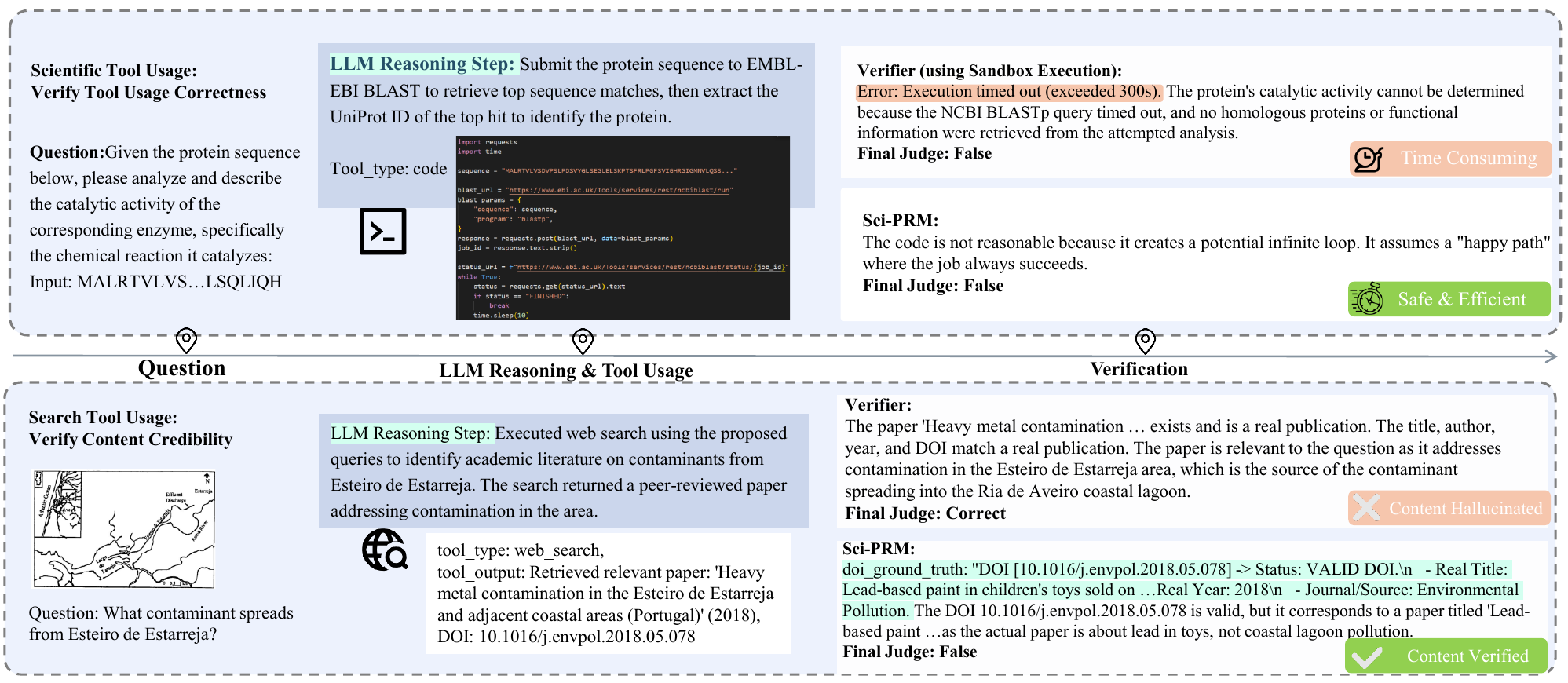}
        \caption{\model{} (Ours) vs. Standard Verifiers. \model{} efficiently detects code logic flaws without execution overhead (top) and accurately identifies citation hallucinations in literature search (bottom).
        }
    \label{fig: case}
 \end{figure*}

Despite these advances, PRM has not been systematically developed for \emph{scientific reasoning}. Scientific reasoning in biology \citep{slake}, chemistry \citep{zhang2024chemllm}, physics\citep{physics}, and earth science \citep{MSEarth, zhou2025omni, feng2025earth} differs fundamentally from math-only reasoning \citep{aime24,deepseekmath, qwen25math}. Correctness often depends not merely on symbolic results, but on evidence-grounded reasoning that must remain consistent with domain knowledge, physical constraints, experimental measurements, and rapidly evolving factual databases. Consequently, LLMs frequently exhibit \textbf{domain-specific hallucinations}: they may produce plausible mechanisms, parameters, or citations that are \textbf{inconsistent} with known science. Importantly, many scientific questions are naturally \emph{tool-verifiable}: a model can access molecular property databases, gene/protein sequence alignment tools, equation solvers, unit converters, or literature retrieval systems to validate intermediate claims. As shown in Figure~\ref{fig:teaser}, general-purpose judges such as GPT-5-Mini maintain reasonable accuracy on ordinary reasoning steps but \emph{degrade obviously} when evaluating trajectories involving scientific tool-calling steps. This gap highlights the need for domain-specialized process judges. However, existing PRM \citep{zou2025reasonfluxprm, wang2025visualprm,zhang-etal-2025-lessons,R-PRM} pipelines rarely incorporate such tools, and most tool-oriented reward modeling works \citep{qian2025toolrl,codeprm, xi2025agentprmprocessrewardmodels,wu2025portooltoolusellmtraining} focus primarily on general domains rather than scientific reasoning. 

At the same time, tool-augmented LLMs \cite{gpt-5, Gemini3, zhao2026openearth, ling2026self, xu2025probing} have emerged as a promising solution for improving reliability. By delegating computation and fact verification to external tools, LLMs can reduce hallucinations and obtain plausible answers. Recent reward modeling studies have begun to explore tool use \citep{qian2025toolrl,codeprm}, demonstrating that specialized reward models can better judge function-calling trajectories than frontier general-purpose LLM judges \citep{zheng2023judging}. Nonetheless, these efforts largely target generic tool environments \citep{toolstar,gou2024critic,toolrm} (e.g., calculators, web search, simple APIs) and emphasize response-level preferences. Scientific reasoning presents \textbf{additional challenges}:  \textbf{(1) choice of tool} must match the scientific subproblem; \textbf{(2) calling} must be correct, including units, formats, and domain-specific parameters; and \textbf{(3) interpretation and utilization} of tool outputs must preserve constraints and uncertainty. A reward model that cannot distinguish these failure modes provides limited signal for training or test-time selection. As illustrated in Figure \ref{fig: case}, our analysis compares \model{} against general judge models regarding their evaluation of common scientific tools, such as web search and code execution. The comparison highlights that scientific domains demand a significantly higher degree of rigor in tool usage. For instance, LLMs utilizing integrated search tools frequently suffer from \textbf{``citation hallucinations''}, generating plausible-sounding but incorrect mappings between paper titles and DOIs—errors that generic verifiers often fail to localize accurately. On the other hand, using code-based tools is time-consuming. Calling scientific APIs (e.g., BLAST for protein alignment or RDKit for molecular processing) often takes 1–5 minutes per run, making them impractical for test-time scaling or as efficient reward signals in reinforcement learning (RL). This underscores the need for a dedicated process reward model like \model{}, which can provide fast, accurate verification to advance scientific LLMs.

We propose \textbf{\model{}}, a \emph{tool-aware process reward model} designed to judge step-by-step scientific reasoning through domain tools. \model{} is trained on a new dataset SCIPRM70K of multi-disciplinary scientific problems paired with reasoning trajectories that contain both a \emph{chain-of-thought} and a \emph{chain-of-tool} (i.e., explicit scientific tool calls, tool outputs, and grounded interpretations). To enable scalable supervision, we introduce an automated step-level labeling pipeline that assigns fine-grained quality labels to each reasoning and tool-use step. In particular, for tool-related steps we label three dimensions: \textbf{(1) tool selection correctness} (is the chosen tool appropriate?), \textbf{(2) tool calling accuracy} (are the arguments and constraints correct?), and \textbf{(3) result validity and utilization} (is the returned evidence correctly interpreted and used downstream?). These labels allow \model{} to produce reliable scalar rewards, capturing where and how scientific reasoning fails.
In the experimental stage, we use \model{} in both the test-time scaling and RL training to validate the effectiveness. During test-time scaling, given a new problem, a base model generates multiple candidate solution trajectories (e.g., Best-of-$N$ sampling). \model{} scores each step and aggregates trajectory-level rewards to select the most reliable path, reducing ``false positives and evidence-inconsistent reasoning''. We test \model{} across representative scientific reasoning benchmarks spanning several disciplines, showing that tool-aware process supervision yields consistent gains over outcome-only judging and over generic tool-use critics, especially on problems requiring multi-step verification. In summary, our contributions are:
\begin{itemize}
    \item \textbf{A tool-augmented scientific dataset SCIPRM70K}. We construct a multi-domain dataset of challenging science questions with trajectories that explicitly interleave reasoning steps and structured tool calls (\emph{chain-of-tool}), enabling supervised learning of evidence-grounded reasoning.
    \item \textbf{Fine-grained step-level supervision for scientific tool use.} We propose an automated labeling framework that distinguishes tool selection, calling, and result utilization errors, providing scalable process supervision beyond math.
    \item \textbf{\model{}: a tool-aware process reward model.} We train a reward model that scores partial scientific reasoning trajectories and supports test-time selection and self-correction, improving reliability and reducing hallucinations.
    \item \textbf{Empirical gains on scientific reasoning.} Across diverse benchmarks, \model{} improves accuracy and factual consistency compared to ORM-style judges and generic tool-use reward models, demonstrating the value of tool-aware process supervision in science.
\end{itemize}

%% file: sections/2_relatedwork.tex
\section{Related Works}

\subsection{Process-Supervised Reward Modeling}
Reward modeling serves as a cornerstone for aligning LLMs with human values and logical consistency \citep{ouyang2022training}. While early approaches primarily relied on Outcome-Supervised Reward Models (ORMs) that evaluate only the final response, recent research has shifted toward Process-Supervised Reward Models (PRMs) \citep{zou2025reasonfluxprm, uesato2022solvingmathwordproblems,zhang-etal-2025-lessons,R-PRM,wang2025visualprm}. PRMs provide fine-grained feedback on intermediate reasoning steps, which has been shown to significantly reduce logical errors and hallucinations compared to ORMs \citep{lightman2024lets, wang2024math}. This granular supervision has proven particularly effective in domains with structured logic, such as mathematics and code generation. However, existing PRM research \citep{qian2025toolrl,codeprm, xi2025agentprmprocessrewardmodels,wu2025portooltoolusellmtraining} focuses predominantly on general domains' reasoning. There is a notable scarcity of research investigating process supervision in specialized scientific domains, such as biomedicine, chemistry, and earth sciences, where reasoning requires not only logical coherence but also rigorous adherence to physical laws and experimental facts.

\subsection{Tool-Augmented Reasoning and Evaluation}
To overcome the limitations of parametric knowledge \cite{wang2025scievalkit, liu2025atlas}, researchers have integrated external tools into LLMs \citep{schick2023toolformer, qin2024toolllm}, enabling capabilities such as web browsing and code execution. In the context of evaluation, recent works have explored ``Tool-assisted Judges'' to enhance the reliability of reward models. For instance, approaches like Math-Shepherd \citep{wang2024math}. While effective for general computational tasks, these approaches typically rely on generic tools (e.g., calculators or search engines) and often employ prompting strategies rather than training the reward model to intrinsically verify tool usage. Recently, autonomous tool construction methods \citep{liu2024apigen} have been proposed. Crucially, existing tool-augmented evaluators lack the specialized capability to validate scientific workflows, such as verifying molecular properties via RDKit or aligning sequences via Biopython. Consequently, they often fail to detect domain-specific errors in parameter selection or the misinterpretation of complex scientific data.

\subsection{Reinforcement Learning for Scientific Reasoning}
Reinforcement learning (RL) has recently been employed to enhance scientific reasoning \cite{zou2026intern, rao2026scidatacopilot, feng2026internagent, tang2025eigen, bai2025intern} and tool-integrated reasoning. Strategies that train models to interleave reasoning with execution have shown promise in improving performance on benchmarks like GSM8K \citep{gsm8k}. However, most existing methods rely on an ``execution-in-the-loop'' paradigm, where the model must execute the tool and receive feedback during the RL training process (e.g., via PPO \citep{ppo} or GRPO \citep{deepseekmath}). While feasible for lightweight tasks like arithmetic, this approach is computationally prohibitive for scientific discovery. Calling specialized scientific packages (e.g., protein folding simulations or complex chemical reaction analysis) entails significant latency, making it impractical to execute these tools for every step during large-scale RL training. Our work addresses this critical bottleneck by proposing \textbf{Sci-PRM}. By serving as a high-quality proxy that scores the reasoning and tool-usage process without requiring live execution during every update, Sci-PRM enables efficient and scalable RL training in complex scientific domains.

%% file: sections/3_method.tex
\section{Creation of the SCIPRM70K Dataset}
\label{sec:datacreat}

\begin{figure*}[t]
	\centering
	\includegraphics[width=\textwidth]{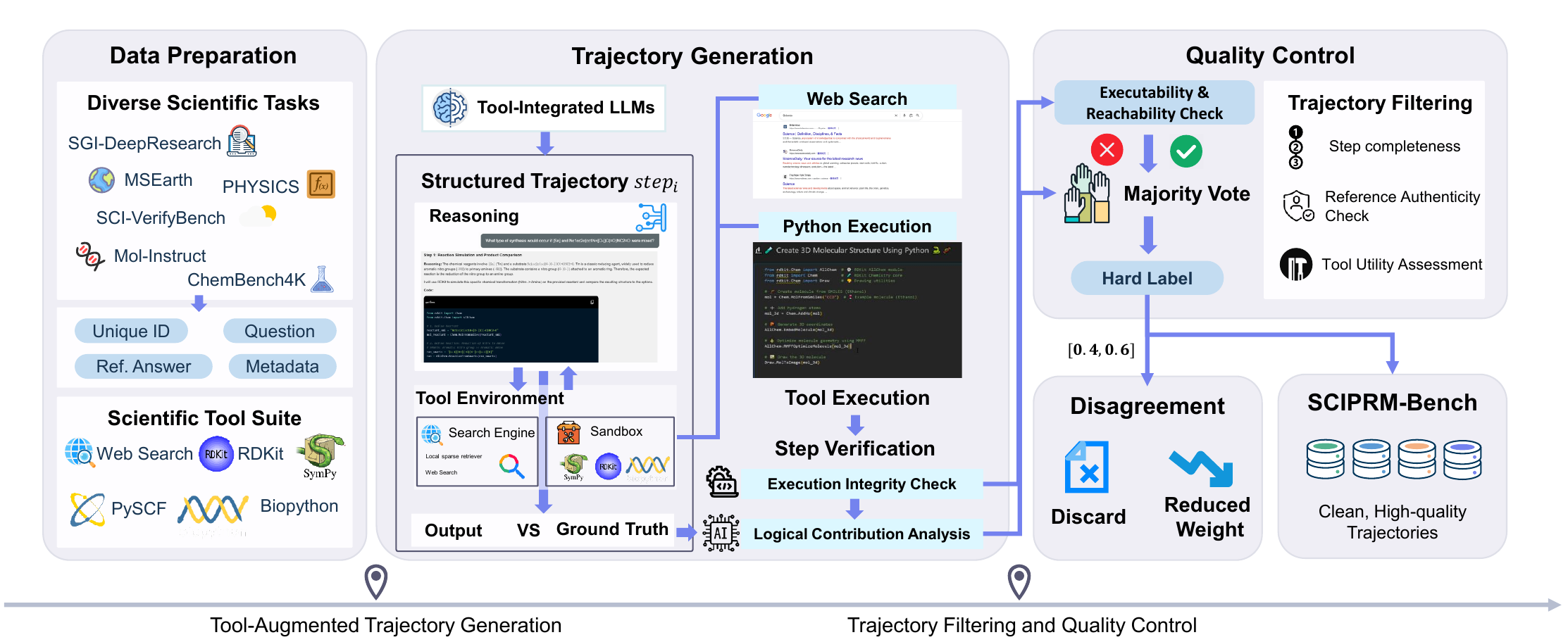}
        \caption{Data curation process for \model{}. The two parts on the left represent data preprocessing, while the two parts on the right encompass the automated generation of VQA and expert-AI collaborative filtering.
        }
    \label{fig: method}
 \end{figure*}

\subsection{Trajectory Preparation}
\label{sec:traj_prep}

\subsubsection{Task Sourcing and Normalization}
\label{sec:task_sourcing}
Most scientific benchmarks evaluate final answers but do not explicitly require tool use~\citep{zheng2025sci, zhang2025compassjudger}. To construct
tool-augmented process data, we collect diverse scientific QA tasks spanning biology, chemistry,
physics, earth science, and general scientific understanding from public sources, including:
general science benchmarks such as \textit{SGI-DeepSearch}~\citep{xu2025probing}, \textit{SCI-VerifyBench}~\citep{zheng2025sci}, and domain-specific benchmarks such as \textit{Mol-Instruction}~\citep{fang2023mol}, \textit{MSEarth}~\citep{MSEarth}, \textit{ChemBench4K}~\citep{zhang2024chemllm}, and \textit{PHYSICS} \citep{physics}. We standardize all tasks into a unified record
\begin{equation}
x = (u, q, a, m),
\end{equation}
where $u$ is a unique id, $q$ is the question, $a$ is the gold answer, and $m$ denotes
optional metadata (e.g., domain, dataset source, and provenance signals such as paper titles or
reference identifiers).

\subsubsection{Tool Suite Construction}
\label{sec:tool_suite}
We support two main tool families:
\begin{itemize}
    \item \textbf{Search tools} for evidence retrieval and citation grounding.
    \item \textbf{Code execution tools} with scientific libraries/APIs for quantitative computation and
    domain-specific processing (e.g., \texttt{rdkit}, \texttt{numpy}, \texttt{sympy}, \texttt{pyscf},
    \texttt{biopython}, etc.).
\end{itemize}
For Code execution tools, we run code in an executable sandbox and record standard outputs
(stdout/return values/exceptions). For search tools, we return top 5 results with source information
(titles/urls/snippets when available). See Table \ref{tab:dataset_stats} for details.

\subsection{Tool-Augmented Trajectory Generation}
\label{sec:gen_verify}

We construct step-labeled trajectories using a \textit{generate--execute--judge} paradigm to obtain reliable step-level supervision.

\subsubsection{Step Generation via Interleaved Execution}
\label{sec:step_gen}
Given each normalized sample $(q,a,m)$, we prompt strong tool-integrated LLMs (e.g.,Gemini-series~\citep{Gemini3} and Doubao-series~\citep{guo2025seed1}) to generate a structured tool-augmented reasoning trajectory. 
Unlike standard Chain-of-Thought reasoning, our generation process is \textbf{interactive}. When the model decides to invoke a tool, the generation stream is temporarily \textit{paused}. The system intercepts the tool call, executes it in the real environment, and appends the result (observation) to the history. The model then \textit{resumes} reasoning based on this actual feedback. This ensures that subsequent reasoning steps are grounded in valid intermediate results rather than hallucinated expectations.

Each step in the trajectory is recorded as:
\begin{align}
z_t = (&\texttt{step\_id},\ \texttt{tool\_used},\ \texttt{tool\_type},\ \texttt{tool\_details},\ \nonumber\\
&\texttt{tool\_output}, \texttt{reasoning\_process}).
\end{align}
A final field \texttt{final\_result} is stored for outcome checking.

\subsubsection{Tool Execution Environment}
\label{sec:execution}
Validating tool steps requires a robust execution backend. During the paused state described above:
\begin{itemize}
    \item \textbf{Web Search:} We issue the query to the search engine and return the top-5 results (snippets/titles/urls) as the observation.
    \item \textbf{Code Execution:} We execute the generated code snippet in a controlled sandbox environment. We capture \texttt{stdout}, return values, and tracebacks (in case of exceptions) as the tool output.
\end{itemize}
This execution feedback closes the loop, enabling the model to self-correct or refine its scientific reasoning based on the results.


\subsection{Trajectory Labeling and Quality Control}
\label{sec:qc}

To train a robust Process Reward Model, we construct a dataset containing both positive reasoning paths and diverse negative samples. We implement a streamlined two-stage pipeline combining execution-based technical validation with MCTS-based logical verification—to assign reliable binary labels $s_t \in \{1,0\}$ to each step.

{Stage 1: Execution-Based Technical Verification.}
This stage assesses the technical feasibility of tool calls and provides grounded observations for subsequent reasoning. For the \texttt{web\_search} tool, the interface returns evidence items containing titles, URLs, and summaries. We assign a negative label ($s_t=0$) to steps resulting in unparsable queries or empty retrieval lists. Similarly, for Python-based tools, we execute generated code within a secure sandbox enforced by a strict 300-second timeout. Execution failures—including syntax errors, runtime exceptions, or timeouts—are automatically labeled as negative. Conversely, valid execution outputs (stdout, return values, or search summaries) are captured and appended to the context. These validated outputs serve as the factual basis for the next stage of logical verification.

{Stage 2: Tool-Aware Step Labeling via MCTS Consistency.}
Steps that pass technical validation may still be logically flawed or irrelevant. Following previous automated function-calling benchmark works~\citep{liu2024apigen}, we leverage LLM as the semantic checker. To distinguish strategically sound steps from subtle reasoning errors, following previous works~\citep{wang2025visualprm,kuang2025tim} we employ Monte Carlo Tree Search (MCTS) consistency checks. We perform $K$ independent rollouts from the current step $z_t$ to the final answer. A step is labeled positive ($s_t=1$) only if it consistently leads to the correct gold answer $a$ exceeding a confidence threshold $\gamma$:
\begin{equation}
P(\text{success} | z_{1:t}) = \frac{1}{K} \sum_{k=1}^K \mathbb{I}(\text{eval}(a_k, a) = \text{True}) \ge \gamma.
\end{equation}
During these rollouts, an LLM judge acts as a critic, evaluating tool selection appropriateness, argument accuracy, and the correct utilization of the tool outputs captured in Stage 1. Steps failing this consistency check are labeled as negative ($s_t=0$). 

\subsubsection{Unified Dataset Construction}
\label{sec:schema}
We aggregate the validated trajectories into a final dataset $D = \{(q, a, m, \{z_t\}_{t=1}^{T}, \{s_t\}_{t=1}^{T})\}$. Each step $z_t$ follows a unified schema:
\begin{align}
z_t = (&\texttt{id},\ \texttt{tool\_used},\ \texttt{tool\_type},\ \texttt{tool\_details},\ \texttt{tool\_output},\nonumber\\
&\texttt{reasoning\_process},\ \texttt{step\_correctness}).
\end{align}
This rigorous labeling process ensures the PRM is trained on grounded scientific inquiry, effectively penalizing both hallucinated tool usage and logical inconsistencies. 

\subsection{Data Statistics}
Utilizing this pipeline, we compile a multi-disciplinary scientific reasoning dataset comprising \textbf{17,818} tool-integrated trajectories and \textbf{86,314} annotated reasoning steps. The data is aggregated from seven high-quality scientific benchmarks covering biology, chemistry, physics, earth science, and general scientific inquiry. Each trajectory contains, on average $\sim4.8$ steps, with domains such as earth science exhibiting longer reasoning chains due to multi-hop verification.

The dataset spans two major tool categories: \textit{Web Search} for factual grounding and \textit{Python Interpreters} for computational analysis (e.g., RDKit molecular computation, physics simulations). We reserve \textbf{1.2k} trajectories as the evaluation set (SCIPRM-Bench), while the remaining data is used for subsequent model training. Figure~\ref{fig:dataset_statistic} and Table~\ref{tab:dataset_stats} provide an overview. See details in Appendix~\ref{dataset_inf}.

\section{Science Process Reward Modeling}

\subsection{Definition}
We present \textbf{\model{}}, a \emph{tool-aware process reward model} for scientific reasoning.
Given a question and a partially generated reasoning trajectory with tool interactions, \model{}
assigns a scalar quality score to the current step conditioned on the full history so far.
This enables (i) \textbf{test-time scaling} via Best-of-$N$ trajectory selection and (ii) \textbf{on-policy RL training} using \model{}
as a reward signal to train tool-using agents with reinforcement learning.

Formally, for each problem instance we represent a reasoning trajectory as a sequence of steps
\begin{equation}
\tau = (z_1, z_2, \ldots, z_T),
\end{equation}
where each step $z_t$ is a structured record containing: whether a tool is used, the tool type, tool-call
details, tool outputs, and the natural-language reasoning for that step. \model{} predicts a score
\begin{equation}
\hat{s}_t = f_{\phi}(q, z_{\le t}) \in [0,1],
\end{equation}
where $z_{\le t}$ denotes the trajectory prefix up to step $t$. Supervision $s_t$ is obtained from
automatic step-level judging (True/False; mapped to $1/0$). We train \model{} with regression
directly on these step labels, without constructing preference pairs.

\noindent\textbf{Overview.} Our pipeline consists of three stages:
(1) constructing a scientific tool-augmented reasoning dataset with explicit Chain-of-Thought and
Chain-of-Tool;
(2) automatically annotating each reasoning step with tool-aware verification to obtain step-level
labels; and
(3) training \model{} and applying it to improve reasoning via test-time scaling and RL.


\begin{figure}[t] 
  \centering
  \includegraphics[width=0.8\textwidth]{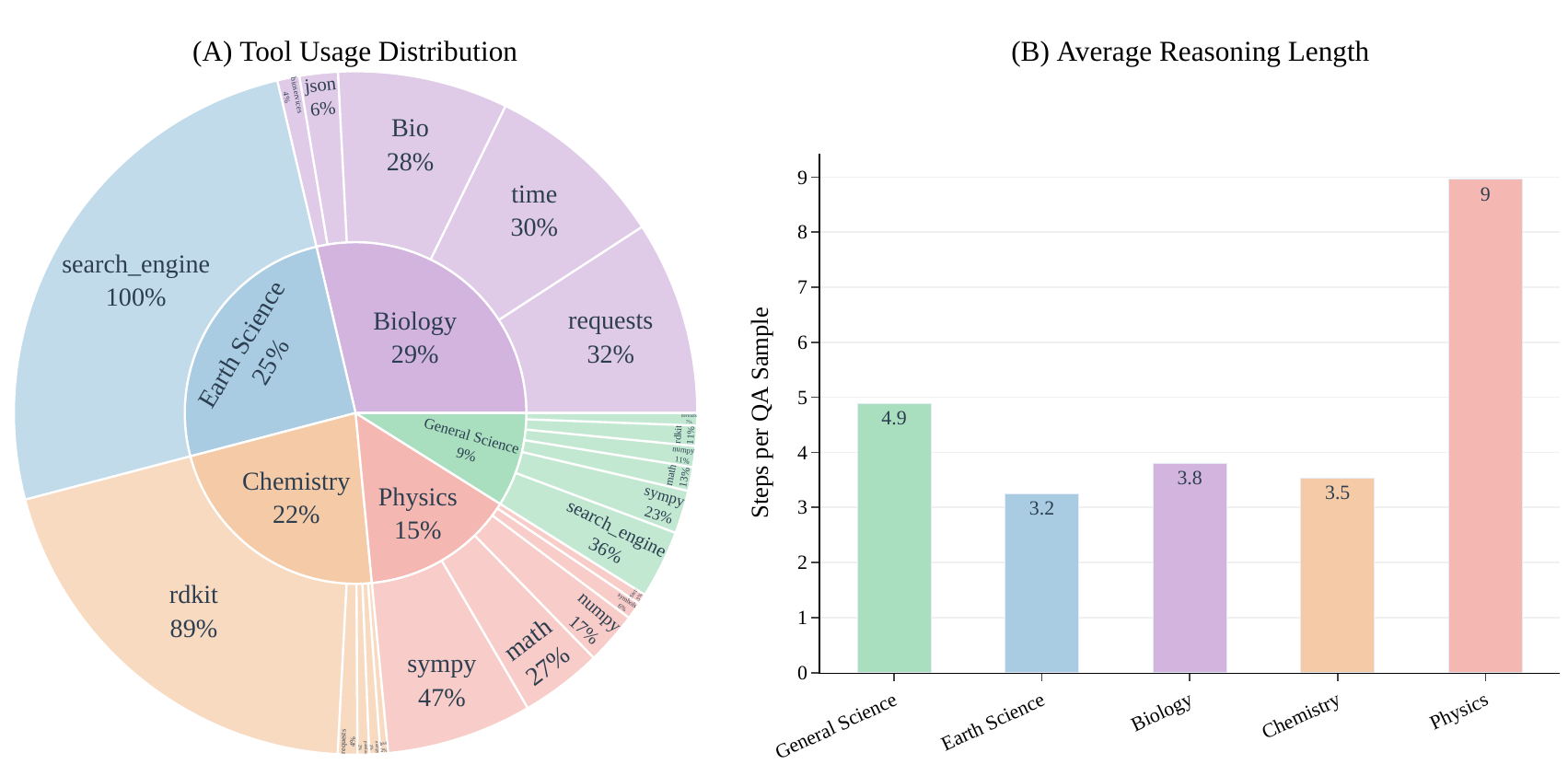}
  \vspace{-2mm}
  \caption{Data statistics of SCIPRM70K.}
  \label{fig:dataset_statistic}
  \vspace{-4mm} 
\end{figure}

\subsection{Dual-Stage Training}
\label{sec:prm_train}

We initialize the \model{} using the {Qwen3-VL-8B} backbone, leveraging its strong multimodal and scientific understanding capabilities. Training follows a two-stage curriculum: domain-specific Supervised Fine-Tuning (SFT) followed by Reinforcement Learning (RL) to enhance verification consistency and generalization.

\subsubsection{Supervised Fine-Tuning (SFT)}
\label{sec:sft}
The first stage focuses on injecting the model with the ability to generate structured reasoning and precise step-level labels. We utilize the curated dataset $D$ described in Sec.~\ref{sec:qc}, which contains high-quality trajectories and their corresponding MCTS-verified labels. Unlike standard SFT that emphasizes output formatting, our verification-centric SFT focuses on transferring domain-specific grounding knowledge. We filter the training traces to retain only concise and valuable reasoning processes, discarding overly verbose or unstructured responses. For each sample, the model is trained to predict the reasoning process and the step-correctness label $s_t$ simultaneously. The training objective is defined as:
\begin{equation}
\mathcal{L}_{\text{SFT}}(\theta) = -\mathbb{E}_{(q, z, s) \sim D} \left[ \log \pi_{\theta}(z_t, s_t \mid q, z_{<t}) \right],
\end{equation}
where $\pi_{\theta}$ represents the model policy. This stage ensures that the model acquires a fundamental understanding of the "stop-and-go" tool-use protocol and the criteria for scientific verification.

\subsubsection{Reinforcement Learning via DAPO-GRPO}
\label{sec:rl_grpo}
To improve the model's robustness and mitigate overfitting on the SFT distribution, we employ \textbf{Dynamic Advantage Policy Optimization (DAPO)} \citep{dapo}, a refined version of Group Relative Policy Optimization (GRPO) \citep{deepseekmath}. This framework allows the model to explore alternative reasoning paths and self-correct based on relative performance within a group of samples.

\paragraph{Dynamic Group Sampling and Filtering.} For each question $q$, we sample a group of $G$ reasoning trajectories $\{o_i\}_{i=1}^G$ from the current policy $\pi_{\theta}$. We implement dynamic sample filtering to stabilize training: trajectories that are trivially easy (where all samples in the group succeed) or excessively difficult (where all fail) are assigned lower weights or filtered out to ensure a meaningful learning signal. To encourage efficiency in scientific inquiry, we incorporate a length penalty into the reward function $R_i$, penalizing redundant tool calls or circular reasoning.

\paragraph{Optimization Objective.} The advantage function $\hat{A}_{i,t}$ is calculated by normalizing the final reward $R_i$ relative to the group mean:
\begin{equation}
\hat{A}_{i,t} = \frac{R_i - \text{mean}(\{R_j\}_{j=1}^G)}{\text{std}(\{R_j\}_{j=1}^G) + \epsilon}.
\end{equation}
The overall objective is to maximize the clipped surrogate reward, ensuring the policy does not deviate too drastically from the old policy $\pi_{\theta_{\text{old}}}$:
\begin{equation}
\begin{split}
\mathcal{J}_{\text{DAPO}}(\theta) = \mathbb{E} \Bigg[ \frac{1}{G} \sum_{i=1}^{G} \sum_{t=1}^{|o_i|} \min \Big( & r_{i,t}(\theta) \hat{A}_{i,t}, \\
& \text{clip}\big(r_{i,t}(\theta), 1-\epsilon, 1+\epsilon\big) \hat{A}_{i,t} \Big) \Bigg],
\end{split}
\end{equation}
where $r_{i,t}(\theta) = \frac{\pi_{\theta}(o_{i,t} | q, o_{i,<t})}{\pi_{\theta_{\text{old}}}(o_{i,t} | q, o_{i,<t})}$ is the importance sampling ratio. This RL stage refines the \model{}'s ability to distinguish nuanced differences in tool usage and evidence interpretation, yielding a more reliable process-based reward signal for scientific tasks.

\subsection{\model{} for Inference and Training}
\label{sec:use_prm}

\subsubsection{Test-Time Scaling via Best-of-$N$ Trajectory Selection}
\label{sec:tts}
At inference time, we prompt a base model to generate $N$ candidate tool-augmented trajectories for a
new question $q$ \citep{snell2025scaling}. \model{} scores each step conditioned on the prefix; we aggregate step scores into a
trajectory score:
\begin{equation}
S(\tau)=\sum_{t=1}^{T} f_{\phi}(q, z_{\le t}) - \lambda \cdot T,
\end{equation}
where $\lambda$ penalizes unnecessarily long trajectories. We select the best trajectory
\begin{equation}
\tau^{\ast} = \arg\max_{\tau_i} S(\tau_i),
\end{equation}
and output its final answer. The base model is prompted with a tool-use template (reason--tool--answer)
to ensure the trajectory is executable and easy to score.

\subsubsection{\model{} as a Reward Model for Tool-Using RL}
\label{sec:rl}
\model{} can further serve as a learned reward signal to train tool-using LLMs/MLLMs with
reinforcement learning. Compared to outcome-only rewards, \model{} provides process-sensitive feedback. In practice, \model{} can be combined with auxiliary rewards such as answer correctness, format adherence, and tool execution success to stabilize training in interactive environments.


%% file: sections/4_exp.tex
\section{Experiments}

\subsection{Datasets}
\label{dataset_details}
\subsubsection{Dataset Setup for Training Sci-PRM}
We train Sci-PRM using the SCIPRM70K corpus described in Section~\ref{sec:datacreat}. 
We adopt a three-way split to support both SFT and RL training.

\noindent\textbf{Training Set.} We allocate \textbf{8.5k trajectories} for SFT, covering diverse scientific domains and tool-use patterns. An additional \textbf{8.1k trajectories} are split into \textbf{27.3K steps} for RL training.

\noindent\textbf{Evaluation Set.} We hold out \textbf{1.5k} trajectories as \textbf{SCIPRM-Bench}, our process-level scientific reasoning benchmark. The evaluation set spans all disciplines and includes both web-search and Python-based tasks, ensuring robust measurement of factual grounding, multi-hop reasoning, and tool-integrated correctness.

\subsubsection{Evaluation Benchmarks for TTS and RL}
We evaluate \model{} from both Test-Time Scaling and RL on 4 representative and challenging scientific benchmarks. Training details can be found in Appendix~\ref{prm_training}. These benchmarks are separate from training data for fair comparison:

 \noindent \textbf{BioProBench~\citep{liu2025bioprobench}:} Focuses on rigorous procedural question answering regarding biological experiments and laboratory protocols.
 
 \noindent \textbf{ChemBench~\citep{mirza2024large}:} Tests advanced chemical reasoning, including reaction prediction and property calculation.
 
 \noindent \textbf{Mol-Instructions (Test Split)~\citep{fang2023mol}:} Assesses the ability to analyze protein structures and biomolecular interactions.
 
\noindent  \textbf{MSEarth (MCQ)~\citep{zhao2025msearth}:} Evaluates earth science reasoning based on observational data and phenomena analysis.

\subsection{Baselines}

We compare our \textbf{\model{}} against various baselines from two distinct perspectives: (1) the judging capability as a reward model \citep{basereward,li2025vlrewardbench}, and (2) the downstream reasoning performance when the reward model is used for inference scaling (Best-of-$N$) \citep{snell2025scaling} and RL training \citep{dapo}.

\noindent\textbf{Multimodal Large Language Models:} We select both strong proprietary and open-source models: GPT-5-Mini \citep{gpt-5}, Gemini-3-Flash \citep{Gemini3}, Llama-3.2-11B-Vision-Instruct \citep{Llama3.2}, GLM-4.6V \citep{zeng2025glm}, Qwen3-VL-8B and Qwen3-VL-32B \citep{qwen3vl}.
    
\noindent \textbf{Test-time Scaling Approches:} We apply different decoding strategies to a fixed base policy model to demonstrate the gains from test-time compute:
\begin{enumerate}
    \item {Majority Voting:} Samples $N$ independent paths and selects the final answer that appears most frequently. This relies on consensus rather than explicit verification.
    \item {ORM-Guided Best-of-$N$:} Samples $N$ paths and selects the one with the highest score according to the Outcome Reward Model described above.
    \item {\model{} Best-of-$N$ (Ours):} Samples $N$ paths and selects the trajectory with the highest aggregated process score, utilizing our fine-grained verification of tool selection, calling, and result interpretation.
\end{enumerate}

\textbf{Reward Model Baselines.}
We compare \model{} against a standard outcome-based approach. \textbf{This model is trained on the same dataset but supervises only the final result}. It treats the entire trajectory as a single unit, assigning a binary label based on the correctness of the final answer (exact matching), while disregarding intermediate tool usage errors or reasoning flaws.

\begin{table*}[t!]
\centering
\caption{Main results on \textbf{SCIPRM-Bench}. We report F1 scores for tool-calling steps across three domains and summarize each model by Global Tool F1, Global No-Tool F1, Overall F1, and the Tool Gap. The Tool Gap is defined as \textit{Global No-Tool F1 $-$ Global Tool F1}, measuring how much a judge degrades when evaluating tool-calling steps. The full Acc/Prec/Rec/F1 table is provided in Appendix.}
\vskip -2mm
\resizebox{\textwidth}{!}{%
\begin{tabular}{ll|cccc|ccc|c}
\toprule
\multirow{2}{*}{\textbf{Model}} & \multirow{2}{*}{\textbf{Type}}
& \multicolumn{4}{c|}{\textbf{Tool-Calling F1} $\uparrow$}
& \multicolumn{3}{c|}{\textbf{Global F1} $\uparrow$}
& \multirow{2}{*}{\textbf{Tool Gap} $\downarrow$} \\
& & ChemBench & Mol-Instruct & Physics & Global
& Tool & No-Tool & Overall & \\
\midrule
\multicolumn{10}{c}{\textit{\textbf{Baseline}}} \\
\midrule
Random & Baseline
& 0.3845 & 0.3343 & 0.3242 & 0.3523
& 0.3523 & 0.6083 & 0.5327 & 0.2560 \\
\midrule
\multicolumn{10}{c}{\textit{\textbf{Closed-Source Models}}} \\
\midrule
Doubao-Seed-1.6 & Closed
& 0.4390 & 0.5092 & 0.8000 & 0.5129
& 0.5129 & 0.7621 & 0.6851 & \textbf{0.2492} \\
GLM-4.6V & Closed
& 0.4711 & 0.4275 & 0.5975 & 0.4689
& 0.4689 & \textbf{0.8540} & 0.7307 & 0.3851 \\
Gemini-3-Flash & Closed
& 0.4771 & \textbf{0.5976} & 0.8144 & 0.5587
& 0.5587 & 0.7860 & 0.7211 & \textbf{0.2273} \\
GPT-5-Mini & Closed
& 0.4630 & 0.4987 & 0.8156 & 0.5338
& 0.5338 & 0.8294 & 0.7592 & 0.2956 \\
\midrule
\multicolumn{10}{c}{\textit{\textbf{Open-Source Models}}} \\
\midrule
Qwen3-VL-8B & Open
& 0.4595 & 0.4792 & 0.7209 & 0.4959
& 0.4959 & 0.7685 & 0.6818 & 0.2726 \\
LLaMA-3.2-11B-V & Open
& 0.1868 & 0.3644 & 0.4624 & 0.3213
& 0.3213 & 0.5754 & 0.5281 & 0.2541 \\
Qwen3-VL-32B & Open
& 0.4273 & 0.4344 & 0.5859 & 0.4569
& 0.4569 & 0.8469 & 0.7539 & 0.3900 \\
\rowcolor{toolbg}
\textbf{Sci-PRM (Ours)} & \textbf{Open}
& \textbf{0.4858} & 0.5362 & \textbf{0.8457} & \textbf{0.5619}
& \textbf{0.5619} & 0.8333 & \textbf{0.7691} & 0.2714 \\
\bottomrule
\end{tabular}%
}
\label{tab:sciprm_bench_main_f1}
\end{table*}

\begin{table}[t]
\centering
\caption{Search Information Verification. It evaluates the models' ability to distinguish between authentic and hallucinated scientific papers retrieved via the \texttt{web\_search} tool. \textbf{HDR}: Hallucination Detection Rate (True Negative Rate).}
\label{tab:search_verification_results}
\vskip 2mm 

\small 
\setlength{\tabcolsep}{12pt} 

\begin{tabular}{l|ccccc}
\toprule
\textbf{Model} & \textbf{Acc.} & \textbf{Prec.} & \textbf{Rec.} & \textbf{F1} & \textbf{HDR} \\
\midrule
Llama-3.2-11B-V & 0.3063 & 0.3264 & 0.6842          & 0.4419          & 0.0529          \\
GLM-4.6V        & 0.4214 & 0.3778 & \textbf{0.7203} & 0.4956          & 0.2265          \\
Doubao-Search   & 0.4333 & 0.3744 & 0.6387          & 0.4720          & 0.2983          \\
Qwen3-VL-32B    & 0.5700 & 0.4342 & 0.2773          & 0.3385          & 0.7624          \\
Qwen3-VL-8B     & 0.6208 & 0.5258 & 0.4322          & 0.4744          & 0.7444          \\
Gemini-3-Flash  & 0.5667 & \textbf{0.7753} & 0.3855 & 0.5149          & 0.8347          \\
GPT-5-Mini      & 0.5932 & 0.4773 & 0.1780          & 0.2593          & 0.8701          \\
\midrule
\rowcolor[HTML]{EFEFEF} 
\textbf{Sci-PRM(Ours)}   & \textbf{0.7458} & \textbf{0.7500} & 0.5385 & \textbf{0.6269} & \textbf{0.8820} \\
\bottomrule
\end{tabular}
\end{table}

\subsection{Evaluation Results on Sci-PRMBench}

\subsubsection{Execution Tools Verification}

Table~\ref{tab:search_verification_results} presents the comprehensive evaluation results on the \model{} Bench, more detailed results can be found in table~\ref{tab:sciprm_bench_main_models} in appendix. A striking observation across all models is the significant performance gap between \textit{Reasoning Capability} (Tool Calling=False) and \textit{Tool Utilization} (Tool Calling=True). While advanced models like GPT-5-Mini and Gemini-3-Flash achieve high F1 scores in reasoning tasks (e.g., 0.8294 and 0.7860 globally), their performance drops precipitously when tasked with verifying scientific code usage (dropping to 0.5338 and 0.5587, respectively). Our model achieves the highest Global F1 score (0.7691), surpassing GPT-5-Mini (0.7592) and significantly outperforming open-source baselines like Qwen3-VL-32B (0.6818).

\textbf{Tool Utilization Competence.} In the critical \textit{Tool Calling=True} split, our model attains the best global F1 score of 0.5619. Notably, in the Physics domain, our model achieves an F1 of 0.8457, demonstrating superior understanding of physical simulation codes compared to Qwen3-VL-32B (0.7209).

\textbf{Reasoning Capability.} Our model maintains competitive reasoning capabilities (F1=0.8333 globally), effectively matching the state-of-the-art GPT-5-Mini (0.8294). This indicates that our training strategy enhances the model’s sensitivity to code correctness without compromising its general scientific reasoning ability, making it a more reliable agent for time-sensitive scientific exploration.

\subsubsection{Search Tool Verification} 

Beyond tool utilization, the capability to verify the authenticity of information retrieved from the web is essential for scientific AI agents. Table~\ref{tab:search_verification_results} summarizes the performance of various models in \textbf{detecting hallucinations} within retrieved academic citations. One can observe that while some baseline models like LLaMA-3.2 and GLM-4.6V exhibit relatively high recall (reaching 0.6842 and 0.7203, respectively), their Hallucination Detection Rate (HDR) is significantly low (e.g., 0.0529 for LLaMA-3.2). This indicates a ``blind trust'' bias, where these models tend to classify most retrieved information as authentic, even when the provided DOIs or titles are fabricated. Conversely, models such as GPT-5-Mini and Gemini-3-Flash demonstrate a more conservative behavior with higher HDR (over 0.83), but suffer from a substantial drop in recall, often misjudging real papers as hallucinations. Our model achieves the best balance between precision and safety. Specifically, Ours outperforms all baseline models in Accuracy (0.7458), F1-score (0.6269), and HDR (0.8820). The high HDR suggests that our model is particularly robust against ``DOI hijacking'' and fabricated metadata, which are common failure modes in automated scientific search. This superior performance is attributed to the integration of rigorous bibliographic verification capabilities, enabling the model to conduct a stringent cross-check between tool outputs and the scientific context. Our experimental results highlight two critical advantages of integrating process-level supervision into scientific MLLMs: the superiority of dense reward signals during training and the reliability of verification during inference.

\begin{figure}[h] 
  \centering
  \includegraphics[width=\linewidth]{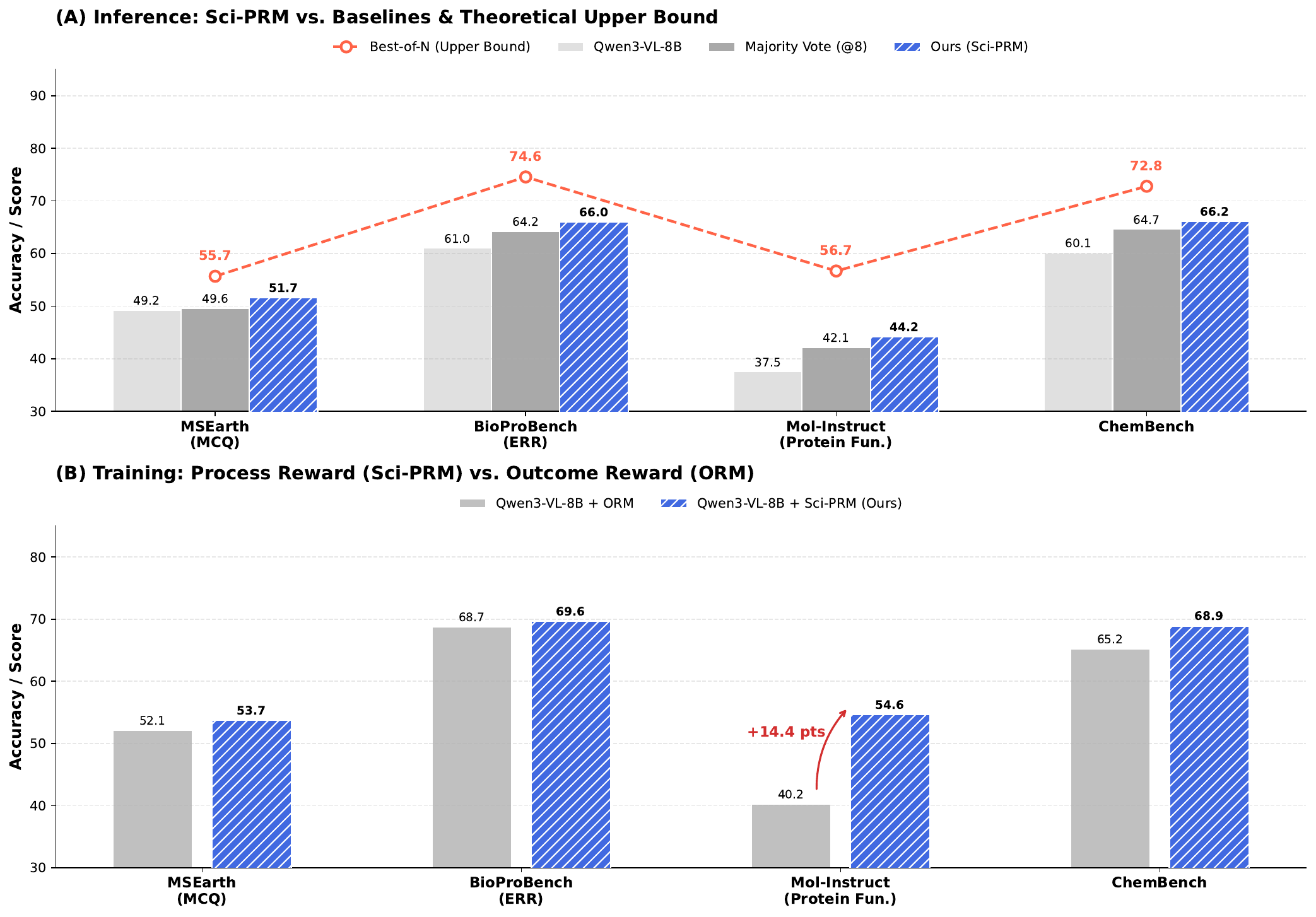}
  
  \caption{\textbf{Efficacy of \model{} in Inference and Training.} 
  (A) Inference: \model{} effectively closes the gap to the Best-of-N upper bound on Qwen3-VL-8B. (B) Training: \model{} demonstrates superior efficiency over ORM, notably gaining +14.4 points on Mol-Instruct.}
  \label{fig:sciprm_ablation}
\vspace{-2mm}
\end{figure}

\begin{table}[htbp]
\centering
\caption{RL Training Results. Comparison of our method against the Outcome Reward Model (ORM) and other representative reward models.}
\label{tab:rl_training_results}
\vskip 2mm
\small
\begin{tabular}{lccc}
\toprule
\textbf{Model} & \textbf{BioProBench-ERR} & \textbf{ChemBench} & \textbf{Mol-Instructions-test} \\
\midrule
ReTool                 & -              & 55.13          & 35.62          \\
Qwen3-VL-8B + ORM      & 68.74          & 65.18          & 40.23          \\
Qwen3-VL-8B + Skywork  & 66.32          & 62.96          & 39.46          \\
\midrule
\rowcolor[HTML]{EFEFEF} 
\textbf{Qwen3-VL-8B + Sci-PRM} & \textbf{69.63} & \textbf{68.90} & \textbf{54.62} \\
\bottomrule
\end{tabular}
\end{table}

\begin{table}[htbp]
\centering
\caption{Test-Time Scaling Results. Evaluation of different verification and scaling strategies.}
\label{tab:test_time_scaling_results}
\vskip 2mm
\small
\begin{tabular}{lccc}
\toprule
\textbf{Model} & \textbf{BioProBench-ERR} & \textbf{Mol-Instructions-test} & \textbf{ChemBench} \\
\midrule
Qwen3-VL-8B                    & 61.04          & 37.53          & 60.13          \\
Qwen3-VL-8B + Majority Vote @8 & 64.21          & 42.14          & 64.65          \\
Qwen3-VL-8B + Skywork          & 64.93          & 41.75          & 64.36          \\
\midrule
\rowcolor[HTML]{EFEFEF} 
\textbf{Qwen3-VL-8B + Sci-PRM} & \textbf{66.04} & \textbf{44.22} & \textbf{66.17} \\
\bottomrule
\end{tabular}
\end{table}




\subsection{Ablation Studies}

\subsubsection{Test-Time Scaling} 
As detailed in Table~\ref{tab:test_time_scaling_results} and supported by Figure~\ref{fig:sciprm_ablation}A, \model{} serves as a robust verifier. We evaluate our approach against standard decoding, \textit{Majority Vote} (@8), and a state-of-the-art general-purpose PRM, Skywork-Reward-Qwen3-8B~\cite{liu2025skywork}. While \textit{Majority Vote} relies on the statistical consensus of redundant samples, it struggles when the base model exhibits systematic hallucinations (i.e., when the majority is wrong). Furthermore, although generic PRMs like Skywork provide token-level verification and improve upon the base model, they lack specialized training on scientific domain reasoning and complex tool-use operations, restricting their effectiveness in scientific scenarios. \model{}, by meticulously evaluating the validity of the scientific reasoning trajectory, effectively filters out plausible-sounding but factually incorrect responses. As shown in the quantitative results, our method surpasses both Majority Vote (@8) and the generic PRM across all domains, and significantly closes the gap towards the empirical \textit{Best-of-N} upper bound, proving that domain-aware verifiable reasoning is a more compute-efficient path to reliability than blind over-generation.

\subsubsection{RL Training} 
In Table~\ref{tab:rl_training_results} and Figure~\ref{fig:sciprm_ablation}B, we validate the effectiveness of \model{} as a reward signal for Reinforcement Learning. We compare \model{} against an Outcome Reward Model (ORM) baseline, as well as two representative reward models: the generic Skywork PRM~\cite{liu2025skywork} and ReTool~\cite{feng2025retool}, a tool-aware reward model. Training with \model{} consistently outperforms all baselines. The advantage of \model{} is particularly pronounced in complex reasoning tasks. Notably, on the \textit{Mol-Instructions} benchmark, which requires multi-step protein function reasoning, \model{} achieves a substantial gain of \textbf{+14.39\%} over the ORM. We attribute this to sparse outcome supervision: ORM only penalizes the final answer, failing to correct intermediate logical fallacies. Moreover, existing generic PRMs and tool-aware RMs struggle to generalize to our setting. Specifically, ReTool is exclusively designed for code sandbox tool interaction and lacks the capability to handle search-based reasoning, rendering it entirely incompatible with the BioProBench evaluation. In contrast, \model{} provides dense, step-aware supervision that comprehends diverse tool invocations (both search and code), guiding the model to follow rigorous scientific principles rather than merely fitting the final answer distribution.

\subsubsection{Efficiency Analysis}
\label{sec:efficiency}

To evaluate the practical deployment feasibility of scientific reasoning models, we analyzed the inference latency and accuracy trade-offs on the \texttt{mol-instruct} subset of Sci-PRMBench (specifically instances requiring Tool Calling). Table~\ref{tab:efficiency} presents the comparison between the base model (\textit{Qwen2.5-VL-7B}), the base model augmented with a Python sandbox for code execution, and our proposed \textit{Sci-PRM}.

\begin{table}[t]
    \centering
    \caption{Efficiency and performance on the SCIPRM-Bench \texttt{mol-instruct} test set (Tool Calling=True). Execution time includes external API latency (e.g., NCBI BLAST).}
    \label{tab:efficiency}
    \begin{tabular}{l c c}
        \toprule
        \textbf{Model} & \textbf{Avg. Time (s)} & \textbf{Accuracy} \\
        \midrule
        Qwen3-VL-8B & 0.72 & 0.51 \\
        Qwen3-VL-8B (w/ Execution) & 0.85 (282.73\textsuperscript{*}) & \textbf{0.86} \\
        \textbf{Sci-PRM (Ours)} & \textbf{0.76} & {0.76} \\
        \bottomrule
    \end{tabular}
\end{table}

As shown in Table~\ref{tab:efficiency}, while augmenting the base model with an execution environment significantly improves accuracy (from 0.51 to 0.81), it introduces a prohibitive latency overhead. The average time skyrockets to 282.73 seconds per query. This delay is primarily driven by external dependencies, such as remote server calls (e.g., BLAST searches) and complex chemical property calculations, rather than the model's inference time itself (which remains low at 0.85s). This magnitude of latency renders the tool-augmented approach impractical for real-time scientific applications.
In contrast, \textit{Sci-PRM} demonstrates a superior balance between efficiency and accuracy. By internalizing scientific reasoning capabilities, our model achieves an accuracy of 0.76—closely approaching the sandbox-augmented upper bound (0.86)—while maintaining an inference speed of 0.76 seconds.
This deficiency in tool utilization is critical in real-world scientific workflows. Scientific tools (e.g., NCBI BLAST, molecular docking simulations) often require substantial computational time—ranging from minutes to hours—to execute. If an agent cannot proactively identify erroneous tool-use code (e.g., incorrect API parameters or logic errors) before execution, it leads to severe inefficiencies and resource wastage. The low precision in the \textit{Tool Calling=True} setting (e.g., Llama-3.2-11b at 0.4954 globally) suggests that current multimodal LLM/MLLMs struggle to distinguish between executable and buggy scientific code, often hallucinating that incorrect code is valid.

%% file: sections/5_conclusion.tex
\section{Conclusion}

We present \model{}, a tool-aware process reward model designed to enforce rigor in scientific reasoning. By supervising step-by-step tool usage—spanning selection, calling, and interpretation—\model{} effectively mitigates hallucinations where general judges and outcome-based models fail. Our empirical results demonstrate that \model{} not only identifies citation fabrications and code logic errors without expensive execution overhead but also significantly boosts downstream performance via inference-time verification. This work underscores the necessity of fine-grained process supervision for building trustworthy and verifiable scientific AI agents.

\section{Acknowledgment}

This work is supported by Shanghai Artificial Intelligence Laboratory.

%% file: sections/6_appendix.tex
\begin{table*}[t]
\centering
\caption{Statistics of the \textbf{\model{}} training dataset after filtering. The dataset covers diverse scientific domains. \textit{Steps} denotes total reasoning steps, while \textit{Tool Usage} counts specific tool callings. The top-5 most frequently used packages are listed for Python-based tasks.}
\label{tab:dataset_stats}
\resizebox{\textwidth}{!}{%
\begin{tabular}{lllccll}
\toprule
\textbf{Discipline} & \textbf{Source Benchmark} & \textbf{Tool Type} & \textbf{QA Samples} & \textbf{Steps} & \textbf{Tool Usage} & \textbf{Top-5 Packages (Freq)} \\ \midrule

\multirow{4}{*}{\textbf{General Science}} 
 & SGI-DeepResearch & Web Search & 318 & 1,350 & 650 & \texttt{search\_engine} (650) \\ \cmidrule(l){2-7} 
 & SCI-VerifyBench & Search / Python & 2,500 & 12,422 & 3,309 & \makecell[l]{\texttt{sympy} (411), \texttt{math} (227) \\ \texttt{numpy} (200), \texttt{rdkit} (193) \\ \texttt{itertools} (118)} \\ \midrule

\textbf{Earth Science} & MSEarth & Web Search & 5,000 & 16,269 & 5,117 & \texttt{search\_engine} (5,117) \\ \midrule

\textbf{Biology} & Mol-Instructions & Python & 5,000 & 18,975 & 8,620 & \makecell[l]{\texttt{requests} (1,848), \texttt{time} (1,732) \\ \texttt{Bio} (1,623), \texttt{json} (365) \\ \texttt{bioservices} (211)} \\ \midrule

\textbf{Chemistry} & ChemBench4K & Python & 4,000 & 14,170 & 4,860 & \makecell[l]{\texttt{rdkit} (4,046), \texttt{requests} (186) \\ \texttt{pandas} (111), \texttt{numpy} (106) \\ \texttt{bs4} (72)} \\ \midrule

\textbf{Physics} & PHYSICS (en) & Python & 1,000 & 8,958 & 2,292 & \makecell[l]{\texttt{sympy} (1,387), \texttt{math} (780) \\ \texttt{numpy} (501), \texttt{symbols} (164) \\ \texttt{Step} (99)} \\ \midrule

\textbf{Total} & \textbf{All Sources} & \textbf{Mixed} & \textbf{17,818} & \textbf{72,144} & \textbf{24,848} & \textbf{-} \\ \bottomrule
\end{tabular}%
}
\end{table*}

\section{Dataset}
\label{dataset_inf}
\subsection{Dataset Details}

We constructed a comprehensive dataset designed to enhance the model's capabilities in scientific reasoning and domain-specific tool utilization. As presented in Table~\ref{tab:dataset_stats}, the curated training dataset encompasses five major scientific disciplines: \textbf{General Science}, \textbf{Earth Science}, \textbf{Biology}, \textbf{Chemistry}, and \textbf{Physics}. The final training corpus consists of 17,818 quality-filtered QA samples, accumulating over 72K steps and nearly 25K specific tool invocations.

The dataset is distinguished by its rich integration of external tools, categorized into Web Search and Python-based code execution.
\begin{itemize}
\item For \textbf{General Science} and \textbf{Earth Science}, the model interacts primarily with search engines (e.g., sources from SGI-DeepResearch and MSEarth) to retrieve up-to-date scientific knowledge.
\item For more specialized domains like \textbf{Biology}, \textbf{Chemistry}, and \textbf{Physics}, the data involves complex Python interactions. As shown in the statistics, the dataset captures high-frequency usage of domain-specific libraries, such as \texttt{rdkit} for chemical structure analysis, \texttt{Bio} (Biopython) for genomic data processing, and \texttt{sympy} for physical equation solving.
\end{itemize}

\paragraph{SCIPRM-Bench Construction.}
To rigorously evaluate the model's performance on unseen data, we constructed a dedicated held-out test set named \textbf{SCIPRM-Bench}. This benchmark was created by randomly selecting a subset of data from four key source benchmarks prior to the training set filtering process. Specifically, the test set composition is as follows:
(1) \textbf{400 samples} from Mol-Instructions (Biology);
(2) \textbf{300 samples} from MSEarth (Earth Science);
(3) \textbf{400 samples} from ChemBench4K (Chemistry); and
(4) \textbf{200 samples} from PHYSICS (Physics).
The remaining data from these sources, along with the General Science samples, constitute the training set detailed in Table~\ref{tab:dataset_stats}.

\subsection{Trajectory Construction}

To construct reasoning paths that leverage the {web\_search} function for solving scientific problems, we employ {Doubao-seed-1.6 (web\_search)} as the path generation model. This model is capable of returning the top-5 search results for a given query, providing the specific title, URL, and summary information for each retrieval. 

Conversely, to construct reasoning paths that utilize Python scientific packages or API functions, we adopt \textbf{Gemini-3-Flash} as the generative model. To ensure the robustness of code execution, we established a sandbox environment pre-installed with a comprehensive suite of scientific packages likely to be utilized during the reasoning process. This isolated environment guarantees that the generated code runs smoothly without interruptions caused by missing dependencies or installation errors. 

\begin{table*}[t!]
\centering
\caption{Results on \textbf{Model} Bench. To highlight the advantages in reasoning, rows for \colorbox{toolbg}{\textbf{Tool Calling=True}} are highlighted with a background color. Models are categorized into Closed-Source and Open-Source.}
\vskip -2mm
\resizebox{\textwidth}{!}{%
\begin{tabular}{ll|cccc|cccc|cccc|cccc}
\toprule
\multirow{2}{*}{\textbf{Model}} & \multirow{2}{*}{\textbf{Split}}
& \multicolumn{4}{c|}{\textbf{ChemBench}}
& \multicolumn{4}{c|}{\textbf{Mol-Instruct}}
& \multicolumn{4}{c|}{\textbf{Physics}}
& \multicolumn{4}{c}{\textbf{Global}} \\
& & Acc & Prec & Rec & F1 & Acc & Prec & Rec & F1 & Acc & Prec & Rec & F1 & Acc & Prec & Rec & F1 \\
\midrule

\multicolumn{18}{c}{\textit{\textbf{Baseline}}} \\
\midrule
\rowcolor{toolbg}
Random
& Tool Calling=True & 0.5000 & 0.3124 & 0.5000 & 0.3845 & 0.5000 & 0.2511 & 0.5000 & 0.3343 & 0.5000 & 0.2399 & 0.5000 & 0.3242 & 0.5000 & 0.2719 & 0.5000 & 0.3523 \\
& Tool Calling=False & 0.5000 & 0.8452 & 0.5000 & 0.6283 & 0.5000 & 0.9429 & 0.5000 & 0.6535 & 0.5000 & 0.5201 & 0.5000 & 0.5098 & 0.5000 & 0.7765 & 0.5000 & 0.6083 \\
& Overall & 0.5000 & 0.6590 & 0.5000 & 0.5686 & 0.5000 & 0.5554 & 0.5000 & 0.5262 & 0.5000 & 0.4296 & 0.5000 & 0.4621 & 0.5000 & 0.5700 & 0.5000 & 0.5327 \\

\midrule
\multicolumn{18}{c}{\textit{\textbf{Closed-Source Models}}} \\
\midrule

\rowcolor{toolbg}
Doubao-Seed-1.6
& Tool Calling=True & 0.3784 & 0.3057 & 0.7784 & 0.4390 & 0.5863 & 0.3626 & 0.8547 & 0.5092 & 0.8847 & 0.6852 & \textbf{0.9610} & 0.8000 & 0.5662 & 0.3691 & 0.8400 & 0.5129 \\
& Tool Calling=False & 0.7085 & 0.9071 & 0.7298 & 0.8088 & 0.4482 & 0.9620 & 0.4318 & 0.5961 & 0.7949 & 0.7536 & 0.9000 & 0.8203 & 0.6718 & 0.8716 & 0.6771 & 0.7621 \\
& Overall & 0.5931 & 0.6750 & 0.7378 & 0.7050 & 0.5255 & 0.5781 & 0.5389 & 0.5578 & 0.8239 & 0.7395 & 0.9110 & 0.8164 & 0.6286 & 0.6629 & 0.7089 & 0.6851 \\
\addlinespace

\rowcolor{toolbg}
GLM-4.6V
& Tool Calling=True & 0.3671 & 0.3188 & \textbf{0.9021} & 0.4711 & 0.3689 & 0.2768 & \textbf{0.9385} & 0.4275 & 0.6978 & 0.4390 & 0.9351 & 0.5975 & 0.4320 & 0.3144 & 0.9222 & 0.4689 \\
& Tool Calling=False & 0.7561 & 0.8773 & 0.8270 & 0.8514 & 0.8679 & 0.9633 & 0.8939 & 0.9273 & 0.7132 & 0.6592 & 0.9286 & 0.7711 & 0.7702 & 0.8430 & 0.8652 & 0.8540 \\
& Overall & 0.6201 & 0.6687 & 0.8395 & 0.7444 & 0.5884 & 0.5834 & 0.9052 & 0.7095 & 0.7082 & 0.6043 & 0.9297 & 0.7325 & 0.6318 & 0.6266 & 0.8764 & 0.7307 \\
\addlinespace

\rowcolor{toolbg}
Gemini-3-Flash
& Tool Calling=True & 0.3929 & 0.3264 & 0.8866 & 0.4771 & 0.7658 & 0.5254 & 0.6927 & \textbf{0.5976} & 0.9034 & \textbf{0.7556} & 0.8831 & 0.8144 & 0.6526 & 0.4267 & 0.8089 & 0.5587 \\
& Tool Calling=False & 0.8382 & 0.9220 & 0.8833 & 0.9022 & 0.2393 & 0.9554 & 0.2027 & 0.3344 & 0.8737 & 0.8691 & 0.8914 & 0.8801 & 0.7078 & 0.9112 & 0.6911 & 0.7860 \\
& Overall & 0.6826 & 0.7075 & 0.8839 & 0.7859 & 0.5342 & 0.6638 & 0.3267 & 0.4379 & 0.8833 & 0.8463 & 0.8899 & 0.8676 & 0.6852 & 0.7283 & 0.7141 & 0.7211 \\
\addlinespace

\rowcolor{toolbg}
GPT-5-Mini
& Tool Calling=True & 0.5443 & 0.3664 & 0.6289 & 0.4630 & 0.7377 & 0.4794 & 0.5196 & 0.4987 & 0.8972 & 0.7157 & 0.9481 & 0.8156 & 0.6961 & 0.4579 & 0.6400 & 0.5338 \\
& Tool Calling=False & 0.7050 & 0.9240 & 0.7093 & 0.8025 & 0.7929 & 0.9558 & 0.8182 & 0.8816 & 0.8009 & 0.7647 & 0.8914 & 0.8232 & 0.7526 & 0.8925 & 0.7747 & 0.8294 \\
& Overall & 0.6488 & 0.7525 & 0.6960 & 0.7232 & 0.7620 & 0.8127 & 0.7426 & 0.7761 & 0.8320 & 0.7549 & 0.9016 & 0.8218 & 0.7295 & 0.7704 & 0.7484 & 0.7592 \\

\midrule
\multicolumn{18}{c}{\textit{\textbf{Open-Source Models}}} \\
\midrule

\rowcolor{toolbg}
Qwen3-VL-8B
& Tool Calling=True & 0.3446 & 0.3095 & 0.8918 & 0.4595 & 0.5091 & 0.3266 & {0.8994} & 0.4792 & 0.8505 & 0.6526 & 0.8052 & 0.7209 & 0.5136 & 0.3452 & \textbf{0.8800} & 0.4959 \\
& Tool Calling=False & 0.7076 & 0.8728 & 0.7656 & 0.8157 & 0.4786 & 0.9436 & 0.4754 & 0.6322 & 0.7578 & 0.7092 & 0.9057 & 0.7955 & 0.6681 & 0.8382 & 0.7094 & 0.7685 \\
& Overall & 0.5808 & 0.6504 & 0.7865 & 0.7120 & 0.4957 & 0.5428 & 0.5827 & 0.5621 & 0.7877 & 0.6993 & 0.8876 & 0.7822 & 0.6048 & 0.6301 & 0.7427 & 0.6818 \\
\addlinespace

\rowcolor{toolbg}
LLaMA-3.2-11B-V
& Tool Calling=True & \textbf{0.6634} & 0.3810 & 0.1237 & 0.1868 & \textbf{0.7896} & 0.7544 & 0.2402 & 0.3644 & 0.7103 & 0.4167 & 0.5195 & 0.4624 & \textbf{0.7269} & \textbf{0.4954} & 0.2378 & 0.3213 \\
& Tool Calling=False & 0.4671 & 0.8425 & 0.4545 & 0.5904 & 0.3982 & 0.9401 & 0.3864 & 0.5477 & 0.5780 & 0.6051 & 0.5429 & 0.5723 & 0.4822 & 0.7921 & 0.4518 & 0.5754 \\
& Overall & 0.5357 & 0.7932 & 0.3997 & 0.5315 & 0.6174 & 0.9015 & 0.3494 & 0.5036 & 0.6207 & 0.5610 & 0.5386 & 0.5496 & 0.5823 & 0.7418 & 0.4100 & 0.5281 \\
\addlinespace

\rowcolor{toolbg}
Qwen3-VL-32B
& Tool Calling=True & 0.4992 & 0.3324 & 0.5979 & 0.4273 & 0.6676 & 0.3792 & 0.5084 & 0.4344 & 0.7445 & 0.4793 & 0.7532 & 0.5859 & 0.6193 & 0.3732 & 0.5889 & 0.4569 \\
& Tool Calling=False & 0.7033 & 0.8507 & 0.7871 & 0.8177 & 0.8946 & 0.9681 & 0.9186 & 0.9427 & 0.7548 & 0.7088 & 0.8971 & 0.7919 & 0.7627 & 0.8485 & 0.8453 & 0.8469 \\
& Overall & 0.6320 & 0.7063 & 0.7558 & 0.7302 & 0.7675 & 0.7773 & 0.8147 & 0.7956 & 0.7515 & 0.6596 & 0.8712 & 0.7508 & 0.7040 & 0.7166 & 0.7952 & 0.7539 \\
\addlinespace

\rowcolor{toolbg}
\textbf{Model (Ours)}
& \textbf{Tool Calling=True}
& {0.5636} & \textbf{0.3844} & {0.6598} & \textbf{0.4858}
& {0.7574} & \textbf{0.5155} & {0.5587} & {0.5362}
& \textbf{0.9159} & {0.7551} & \textbf{0.9610} & \textbf{0.8457}
& {0.7154} & {0.4832} & {0.6711} & \textbf{0.5619} \\
& Tool Calling=False
& 0.7102 & 0.9246 & 0.7155 & 0.8067
& 0.7982 & 0.9560 & 0.8239 & 0.8850
& 0.8053 & 0.7690 & 0.8943 & 0.8269
& 0.7576 & 0.8943 & 0.7801 & 0.8333 \\
& Overall
& 0.6590 & 0.7594 & 0.7062 & 0.7319
& 0.7753 & 0.8243 & 0.7567 & 0.7891
& 0.8410 & 0.7663 & 0.9063 & 0.8305
& 0.7404 & 0.7798 & 0.7588 & \textbf{0.7691} \\

\bottomrule
\end{tabular}%
}
\label{tab:sciprm_bench_main_models}
\end{table*}

\section{Implementation Details}

All the experiments are conducted on two NVIDIA H200 GPUs. We use Qwen3-VL-8B-Instruct as the base model and employ the ms-swift framework for training. We adopt Low-Rank Adaptation (LoRA) for parameter-efficient fine-tuning across both stages. First, we perform Supervised Fine-Tuning (SFT) for 1 epoch with a learning rate of $1 \times 10^{-4}$ and a warmup ratio of 0.05. Subsequently, we align the model using DAPO with a learning rate of $5 \times 10^{-7}$, a group size of 8 generations, and a KL coefficient $\beta$ of 0.001. As for the fine-grained beam search with \model{}, we set the temperature as 0.8, and select the number of beams $N$ and the beam width $M$ from $\{1, 2, 4, 8\}$. The detailed hyperparameters for SFT and RL phases are listed in Table~\ref{tab:sft_config} and Table~\ref{tab:rl_config}, respectively.

\begin{table}[h]
    \centering
    \caption{SFT Configurations for SCI-Verifier.}
    \label{tab:sft_config}
    \begin{tabular}{lc}
        \toprule
        \textbf{Parameter} & \textbf{Value} \\
        \midrule
        Base Model & Qwen3-VL-8B-Instruct \\
        Precision & BF16 \\
        Training Framework & ms-swift \\
        Training Type & LoRA \\
        LoRA Rank / Alpha & 8 / 32 \\
        Target Modules & all-linear \\
        Optimization Strategy & DeepSpeed Zero2 \\
        Learning Rate & $1 \times 10^{-4}$ \\
        Warmup Ratio & 0.05 \\
        Max Sequence Length & 16384 \\
        Number of Training Epochs & 1 \\
        Per Device Train Batch Size & 2 \\
        Gradient Accumulation Steps & 2 \\
        GPUs Per Node & 1 \\
        \bottomrule
    \end{tabular}
\end{table}

\begin{table}[h]
    \centering
    \caption{RL (DAPO) Configurations for SCI-Verifier.}
    \label{tab:rl_config}
    \begin{tabular}{lc}
        \toprule
        \textbf{Parameter} & \textbf{Value} \\
        \midrule
        RL Algorithm & DAPO \\
        Precision & BF16 \\
        Training Type & LoRA (Rank 8, Alpha 32) \\
        Optimization Strategy & DeepSpeed Zero2 \\
        Learning Rate & $5 \times 10^{-7}$ \\
        Warmup Ratio & 0.01 \\
        KL Coefficient ($\beta$) & 0.001 \\
        Temperature & 1.0 \\
        Number of Generations & 8 \\
        Max Prompt Length & 4096 \\
        Max Completion Length & 4096 \\
        Number of Training Epochs & 1 \\
        Number of Iterations & 1 \\
        Per Device Train Batch Size & 2 \\
        Gradient Accumulation Steps & 16 \\
        Reward Function & verdicts\_acc, soft\_overlong \\
        \bottomrule
    \end{tabular}
\end{table}

\section{PRM \& ORM Training Details}
\label{prm_training}
To validate the effectiveness of Sci-PRM as a reward model within the Reinforcement Learning framework, we conduct comparative experiments based on the Qwen3-VL-8B model. We employ Group Relative Policy Optimization (GRPO) as the training algorithm. To assess the impact of dense process supervision, we compare two distinct reward configurations:
\begin{itemize}
\item \textbf{ORM-based GRPO:} The model is trained using an Outcome Reward Model (ORM) that provides sparse feedback based solely on the correctness of the final answer.
\item \textbf{Sci-PRM-based GRPO:} The model is trained using our proposed Sci-PRM, which provides fine-grained, step-wise reward signals throughout the reasoning process.
\end{itemize}

The training dataset is constructed to cover diverse scientific tasks, sourced from the MSEarth training dataset, Mol-Instruct training dataset, BioProtein training dataset and ChemBench4K. Detailed hyperparameters for the GRPO training phase are provided in Table~\ref{tab:prmrl_config}. The quantitative results are summarized in Table~\ref{tab:rl_training_comparison}. As illustrated, Reinforcement Learning significantly enhances the model's performance compared to the supervised fine-tuning (SFT) baseline (Qwen3-VL-8B). While the ORM-based GRPO yields consistent improvements across all domains (e.g., improving from 60.13 to 65.18 on ChemBench and 61.04 to 68.74 on BioProBench), the Sci-PRM-based approach demonstrates superior performance on all benchmarks. Notably, the advantage of dense process supervision is most pronounced in complex reasoning tasks. On the \textbf{Mol-Instructions} benchmark, which requires intricate biomolecular reasoning, the ORM method only achieves a modest gain (40.23 vs. 37.53). In contrast, our Sci-PRM achieves a remarkable score of \textbf{54.62}, outperforming the ORM baseline by over 14 points. This substantial margin indicates that for tasks involving long chains of thought, sparse outcome signals are insufficient, whereas fine-grained step-wise rewards provided by Sci-PRM successfully guide the model toward correct reasoning paths.

\begin{table}[h]
    \centering
    \caption{RL (GRPO) Configurations.}
    \label{tab:prmrl_config}
    \begin{tabular}{lc}
        \toprule
        \textbf{Parameter} & \textbf{Value} \\
        \midrule
        RL Algorithm & GRPO \\
        Precision & BF16 \\
        Training Type & LoRA (Rank 8, Alpha 32) \\
        Optimization Strategy & DeepSpeed Zero2 \\
        Learning Rate & $5 \times 10^{-7}$ \\
        Warmup Ratio & 0.01 \\
        KL Coefficient ($\beta$) & 0.001 \\
        Temperature & 1.0 \\
        Number of Generations & 8 \\
        Max Prompt Length & 4096 \\
        Max Completion Length & 4096 \\
        Number of Training Epochs & 1 \\
        Number of Iterations & 1 \\
        Per Device Train Batch Size & 2 \\
        Gradient Accumulation Steps & 16 \\
        Reward Function & verdicts\_acc, soft\_overlong \\
        \bottomrule
    \end{tabular}
\end{table}

\begin{table}[htbp]
    \centering
    \caption{Performance comparison on scientific benchmarks.}
    \label{tab:sci_benchmarks_final}
        \setlength{\tabcolsep}{6pt} 
        \renewcommand{\arraystretch}{1.25} 
        \begin{tabular}{lcccc}
            \toprule
            \textbf{Benchmark} & \textbf{Qwen3-VL} & \textbf{MV@8} & \textbf{+Sci-PRM} & \textbf{Best-of-N} \\
            \midrule
            Msearth & 49.21 & 49.57 & 51.66 & \textbf{55.68} \\
            BioProBench & 61.04 & 64.21 & 66.04 & \textbf{74.56} \\
            Mol-Instructions & 37.53 & 42.14 & 44.22 & \textbf{56.68} \\
            ChemBench & 60.13 & 64.65 & 66.17 & \textbf{72.79} \\
            \bottomrule
        \end{tabular}
    
    \vspace{1ex}
    \footnotesize
    \raggedright
    \textit{Note:} MV@8: Majority Vote (8 samples). The proprietary model (Mol-Instructions) achieves a Rouge-L score of 52.00. Bold values denote the best performance.
\end{table}

\begin{table}[htbp]
    \centering
    \caption{Performance comparison of RL training methods. We compare the base model (SFT) against RL training with ORM and our Sci-PRM.}
    \label{tab:rl_training_comparison}
    \renewcommand{\arraystretch}{1.25} 
    
        \begin{tabular}{lccc}
            \toprule
            \multirow{2.5}{*}{\textbf{Benchmark}} & \textbf{Base Model} & \multicolumn{2}{c}{\textbf{RL Training}} \\
            \cmidrule(l){3-4} 
             & (Qwen3-VL) & \textbf{+ORM} & \textbf{+Sci-PRM (Ours)} \\
            \midrule
            Msearth & 49.21 & 52.12 & \textbf{53.72} \\
            BioProBench-ERR & 61.04 & 68.74 & \textbf{69.63} \\
            ChemBench & 60.13 & 65.18 & \textbf{68.90} \\
            Mol-Instructions & 37.53 & 40.23 & \textbf{54.62} \\
            \bottomrule
        \end{tabular}
    
    \vspace{1ex}
    \footnotesize
    \raggedright
    \textit{Note:} All models are initialized from Qwen3-VL-8B. The \textbf{+ORM} baseline uses a standard Outcome Reward Model, while \textbf{+Sci-PRM} utilizes our fine-grained process reward model. Bold values indicate the best performance.
\end{table}

\section{Model}

We evaluate a diverse set of large language models (LLMs) covering both general-purpose and tool-aware reasoning systems:

\begin{itemize}
  \item \textbf{Doubao-Seed-1.6}: A general-purpose instruction-following LLM deployed via an API service. We use it as a strong industrial baseline for structured-output generation and tool-aware reasoning, since it generally supports stable chat-style prompting and deterministic decoding settings.

  \item \textbf{Gemini-3-Flash}: A latency- and throughput-optimized model designed for fast response. We include it to examine how an efficiency-oriented model behaves on verification tasks, especially for enforcing strict output formats and maintaining consistent reasoning across multi-step prompts.

  \item \textbf{Qwen3-VL-8B}: A compact model from the Qwen ``VL'' family (8B scale). It serves as a smaller-capacity baseline to characterize the effect of limited model scale on (i) compliance with step schemas, (ii) detection of flawed reasoning/tool misuse, and (iii) overall verification accuracy.

  \item \textbf{LLaMA-3.2-11B-Vision-Instruct}: An instruction-tuned model from the Llama family with vision-language capabilities. Although our evaluation focuses on text-based verification prompts, this model is included for breadth and to assess how a general instruction-following model (with multimodal pretraining) performs on structured, tool-aware reasoning-chain verification.

  \item \textbf{Qwen3-VL-32B}: A larger-capacity model in the Qwen ``VL'' family (32B scale). We use it as a higher-capacity open-model baseline to quantify scaling effects relative to \textbf{Qwen3-VL-8B}, and to test whether increased capacity improves the detection of subtle logical errors and inconsistent tool usage.

  \item \textbf{gpt-5-mini}: A compact GPT-family model used as a high-quality commercial baseline. We include it to benchmark verification performance under strong instruction-following behavior, particularly on strict schema compliance, tool-specification correctness (query/code/API usage), and robustness across different splits.

\end{itemize}

\section{SCIPRM Reasoning-Chain Dataset and Prompts}

\subsection{Dataset Construction Details}

\begin{promptbox}{Biomolecular (Protein \& Molecular) Data}
\begingroup
\small\ttfamily\raggedright\sloppy
\setlength{\parindent}{0pt}
\setlength{\parskip}{0.35em}

You are a helpful assistant and a professional bioinformatics researcher.
You are in an English-speaking environment. Please answer strictly in English.

\textbf{Question:} \{question\}

\textbf{Input Protein Sequence:}
\{sequence\}

Answer this question. You can use tools; if you use tools, you must provide specific code.
I will run the code and give you the output for your next step of reasoning.
The code needs to be directly runnable and concise. Do not require manual API key insertion if possible.

\textbf{OUTPUT FORMAT (STRICT):}
Return STRICT JSON only (no extra text).
The root MUST be a JSON ARRAY; each element is one reasoning step.
Each step MUST include:
- step\_id: number
- tool\_used: boolean
- tool\_type: string ("python\_code"\allowbreak{} / "scientific\_api"\allowbreak{} / "none")
- tool\_details: string (runnable code\allowbreak{} / exact query;\allowbreak{} empty string if tool\_used is false)
- reasoning: string (what the step checks\allowbreak{} and how it supports discriminating among options;\allowbreak{} do not invent tool results)

\endgroup
\end{promptbox}

\begin{promptbox}{Physical Sciences Data}
\begingroup
\small\ttfamily\raggedright\sloppy
\setlength{\parindent}{0pt}
\setlength{\parskip}{0.35em}

You are a rigorous physics and scientific researcher.
Given a physics problem, produce an evidence-supported reasoning chain to solve it.
You may propose tool steps (especially runnable Python code for calculation),
but you cannot execute tools yourself.
If you need computation, consolidate everything into fewer Python code blocks.
All other steps must have tool\_type 'none'.

Question:
\{question\}

I need an evidence-supported reasoning chain to solve this problem.
You may use tools such as Python packages to support the calculation and you need to give me the final results.
If you use tools, you must provide specific code.
The code needs to be directly runnable and concise. Do not require manual API key insertion if possible.

TOOL RULES (STRICT):
- If a tool is needed, put ONLY directly runnable code (or an exact query string) in tool\_details.
- Do NOT include any explanation, comments, markdown, or extra text inside tool\_details.

PYTHON\_CODE RULES:
- If tool\_type is "python\_code", tool\_details MUST be a single runnable Python snippet.
- The code MUST end with at least one print(...)\allowbreak{} that prints the final computed evidence (e.g., a number or dict).
- Do not rely on implicit/interactive display of variables.

OUTPUT FORMAT (STRICT):
Return STRICT JSON only (no extra text).
The root MUST be a JSON ARRAY; each element is one reasoning step.
Each step MUST include:
- step\_id: number
- tool\_used: boolean
- tool\_type: string ("python\_code"\allowbreak{} / "scientific\_api"\allowbreak{} / "none")
- tool\_details: string (runnable code\allowbreak{} / exact query;\allowbreak{} empty string if tool\_used is false)
- reasoning: string (what the step checks/calculates\allowbreak{} and how it contributes to the solution)

\endgroup
\end{promptbox}

\begin{promptbox}{Chemical Sciences Data}
\begingroup
\small\ttfamily\raggedright\sloppy
\setlength{\parindent}{0pt}
\setlength{\parskip}{0.35em}

\textbf{SYSTEM PROMPT:}
You are a rigorous chemistry/cheminformatics researcher.
Given a multiple-choice question asking which compounds are commonly used to synthesize a target SMILES,
you must produce an evidence-supported reasoning chain.
You are allowed to propose tool steps (especially runnable Python/RDKit code),
but you cannot execute tools yourself.

\textbf{USER PROMPT TEMPLATE:}
Question:
\{question\}

Options:
A: \{A\}
B: \{B\}
C: \{C\}
D: \{D\}

I need an evidence-supported reasoning chain.
You may use tools such as Python packages (e.g., RDKit) to support the reasoning.
If you use tools, you must provide specific code.
The code needs to be directly runnable and concise.
Do not require manual API key insertion if possible.
I will run the code and provide the outputs to you later;\allowbreak{} therefore, do NOT fabricate tool outputs.

IMPORTANT tool rule:
- If a tool is needed, put ONLY directly runnable code (or an exact query string) in tool\_details.
- Do NOT include any explanation, comments, markdown, or extra text inside tool\_details.

Output requirement:
Output the answer in STRICT JSON format.
The ROOT must be a JSON ARRAY, where each element is one reasoning step.
Each step MUST include the following fields:
- tool\_used: boolean (true/false)
- tool\_type: string ("python\_code"\allowbreak{} / "scientific\_api"\allowbreak{} / "none")
- tool\_details: string (runnable code;\allowbreak{} or empty string if tool\_used is false)
- reasoning: string (explain what this step is verifying)

Additional constraints:
- Prefer python\_code steps that verify structural compatibility\allowbreak{} (substructure matching, key functional-group motifs, molecular-weight sanity checks).
- If you can reasonably decide without tools, you may set tool\_used=false,\allowbreak{} but you should still provide evidence-based chemical logic.
- Do NOT output the final choice separately;\allowbreak{} the chain should make the decision clear in the reasoning steps.

Return STRICT JSON only (no extra text).

\endgroup
\end{promptbox}

\begin{promptbox}{Earth \& Environmental Sciences Data}
\begingroup
\small\ttfamily\raggedright\sloppy
\setlength{\parindent}{0pt}
\setlength{\parskip}{0.35em}

\textbf{SYSTEM PROMPT:}
You are a careful clinical/scientific reasoning assistant.
You must produce STRICT JSON only (no extra text).
If you use any tool, you must describe exact usage\allowbreak{} (query string\allowbreak{} / runnable python code\allowbreak{} / API endpoint).

\textbf{USER PROMPT (ANSWER TEMPLATE):}
Question: \{question\}

You can use tools to help you answer\allowbreak{} (web\_search\allowbreak{} / python\_code\allowbreak{} / scientific\_api\allowbreak{} / none).
Return the answer step by step in English.

Output STRICT JSON with this schema:
\{
  "steps": [
    \{
      "step\_id": 1,
      "tool\_used": true/false,
      "tool\_type": "web\_search"\allowbreak{}|"python\_code"\allowbreak{}|"scientific\_api"\allowbreak{}|"none",
      "tool\_details": "exact query string OR full runnable python code OR exact API usage OR 'none'",
      "tool\_output": "retrieved info / code execution result / api response summary / 'none'",
      "reasoning\_process": "explain the reasoning for this step"
    \}
  ],
  "final\_result": "final answer to the question"
\}

Constraints:
- If you choose web\_search,\allowbreak{} include the exact query string.
- If you choose python\_code,\allowbreak{} include complete runnable code\allowbreak{} (no placeholders).
- If you do not use tools in a step,\allowbreak{} set tool\_type="none",\allowbreak{} tool\_details="none",\allowbreak{} tool\_output="none".
Return STRICT JSON only.

Constraints:
- Tool usage requirements are the same:\allowbreak{} exact query\allowbreak{} / runnable python\allowbreak{} / explicit API usage.
- step\_correctness should assess the logic/content of that step\allowbreak{} (not tool availability).
Return STRICT JSON only.

\endgroup
\end{promptbox}

\subsection{Evaluation Prompts}

\begin{promptbox}{Evaluation Prompts (Tool-use Verifiers)}
\begingroup
\small\ttfamily\raggedright\sloppy
\setlength{\parindent}{0pt}
\setlength{\parskip}{0.35em}

\textbf{A.\ Evaluation: Questions Requiring Code Execution or API Tools}

\textbf{SYSTEM PROMPT:}
You are an expert process verifier.
Your task is to evaluate the proposed reasoning step and tool-usage code BEFORE execution.
Check if the logic is sound and the code is correct for the given problem.
The execution result of the previous step\allowbreak{} (if any) is provided as context.
Response format:\allowbreak{} "Valid" or "Invalid" only.

\textbf{USER PROMPT TEMPLATE (per step):}
Question:
\{question\}

Observation (Output from previous step)\allowbreak{} [optional]:
\{previous\_tool\_output\}

Current Step \{step\_id\}:
Reasoning: \{reasoning\}

Tool Type: \{tool\_type\}\allowbreak{} [only if tool\_used=true]
Proposed Code:\allowbreak{} [only if tool\_used=true]
\{tool\_details\}

Is this step valid and correct?

\textbf{ASSISTANT OUTPUT FORMAT (STRICT):}
Return exactly one token:\allowbreak{} "Valid" or "Invalid".
No extra text.

\medskip
\textbf{B.\ Evaluation: Prompts Requiring Retrieval / Search Tools}

\textbf{SYSTEM PROMPT:}
You are an expert scientific fact-checker and researcher.
Your task is to verify the authenticity and relevance of a specific academic paper/citation retrieved by an AI agent during a problem-solving process.
Strictly output your response in JSON format.

\textbf{USER PROMPT TEMPLATE (per step to verify):}
\textbf{Problem Context}
Context Paper Title:\allowbreak{} \{context\_title\}
Question:\allowbreak{} \{question\}
Image Caption:\allowbreak{} \{caption\}

\textbf{Reasoning History}
\{history\}\allowbreak{} \,\, (or "No previous steps.")

\textbf{Step to Verify (Web Search / Paper Retrieval)}
Tool Used:\allowbreak{} \{tool\_type\}
Tool Input (Query):\allowbreak{} \{tool\_input\}
Tool Output (Retrieved Paper/Info):\allowbreak{}
\{tool\_output\}

Reasoning:\allowbreak{} \{reasoning\}

\textbf{Your Verification Tasks:}
1.\allowbreak{} \textbf{Authenticity Check (Crucial):}
- Does the paper mentioned in the Tool Output actually exist?
- Do the Title, Author (if any), Year, and DOI (if have) match a real publication?
- Important:\allowbreak{} If the DOI is fake, or if the title does not belong to the DOI, mark it as Hallucinated.

2.\allowbreak{} \textbf{Relevance Check:}
- If the paper exists, is it helpful for answering the specific Question above?

\textbf{OUTPUT FORMAT (STRICT JSON):}
\{
  "status": "Authentic"\allowbreak{} or "Hallucinated",
  "analysis": "Step-by-step verification logic.\allowbreak{} First state if the DOI/Title exists.\allowbreak{} Then state if it is relevant."
\}

\textbf{NOTES:}
- Return "Authentic" ONLY if the paper is real AND the metadata\allowbreak{} (DOI/Year if have)\allowbreak{} is consistent.
- Return "Hallucinated" if the paper does not exist,\allowbreak{} the DOI is fake,\allowbreak{} or the title/DOI mismatch.

\endgroup
\end{promptbox}

\section{Case Study}

\subsection{Case 1: Code Execution Verification}
This case demonstrates the verification of a Python script interacting with the InterProScan API. The proposed code contains logic errors (incorrect header usage and return parsing).

\begin{promptbox}{Input: Protein Analysis Request}
\textbf{Question:} Analyze the protein corresponding to the amino acid sequence: \texttt{MSDSEQT...} (truncated).

\textbf{Proposed Step:} Submit sequence to InterProScan via EBI REST API to identify domains.
\textbf{Tool Type:} \texttt{python\_code}

\textbf{Proposed Code:}
\begin{lstlisting}[language=Python]
import requests
def submit_interproscan(sequence, email):
 url = "https://www.ebi.ac.uk/Tools/services/rest/interproscan/run"
 headers = {
  "Content-Type": "application/x-www-form-urlencoded",
  "Accept": "application/json",
  "X-EMAIL": email # Incorrect placement
 }
 data = {
  "sequence": sequence,
  "type": "protein",
  "tool": "interproscan",
  "stype": "sequence"
 }
 response = requests.post(url, headers=headers, data=data)
 if response.status_code == 200:
  return response.json() # Incorrect response parsing
 else:
  return f"Error: {response.status_code}"

seq = "MSDSEQT..."
result = submit_interproscan(seq, "user@example.com")
print("Job ID:", result.get('jobId'))
\end{lstlisting}

\textbf{Actual Execution Result (Stderr):}
\texttt{AttributeError: 'str' object has no attribute 'get'}
\end{promptbox}

\begin{promptbox}{Model Evaluations}
\textbf{Ours:} \statusbadge{Invalid} \\
\textit{Reason:} The code incorrectly passes 'email' in headers instead of body, uses invalid 'stype', and attempts to parse plain-text Job ID as JSON.

\vspace{1mm}\hrule\vspace{1mm}

\textbf{Qwen3-VL-8B:} \statusbadge{Valid} \\
\textit{Reason:} The endpoint URL and headers are logically aligned... no runtime errors are evident in the logic.

\vspace{1mm}\hrule\vspace{1mm}

\textbf{GPT-5-Mini:} \statusbadge{Invalid} \\
\textit{Reason:} The EBI Tools endpoint expects form fields like "email" and typically returns a plain-text job identifier... relying on \texttt{response.json()} will fail.
\end{promptbox}

\subsection{Case 2: Search Result Verification}

This example illustrates the verification of a retrieved citation regarding the "Southern Annular Mode."

\begin{promptbox}{Input: Search Verification}
    \textbf{Question:} Which meteorological phenomenon is linked to stronger zonal winds in this figure? (Figure context provided).
    
    \textbf{Retrieved Item:}
    \begin{itemize}
        \item \textbf{Title:} The Southern Annular Mode and its influence on Southern Hemisphere climate
        \item \textbf{DOI:} 10.1029/1999JD901133
        \item \textbf{Year:} 2000
    \end{itemize}
    
    \vspace{2mm}
    \textbf{DOI Ground Truth:}
    \begin{itemize}
        \item \textbf{DOI:} 10.1029/1999JD901133 $\rightarrow$ Status: \textbf{VALID DOI}
        \item \textbf{Real Title:} Simulations of water vapor in the lower stratosphere and upper troposphere
        \item \textbf{Real Authors:} Andrew Gettelman, James R. Holton, Anne R. Douglass
        \item \textbf{Real Year:} 2000
        \item \textbf{Journal/Source:} Journal of Geophysical Research: Atmospheres
    \end{itemize}
\end{promptbox}

\begin{promptbox}{Model Evaluations}
    \textbf{Ours:} \textcolor{success_green}{\textbf{[Hallucinated]}} \\
    \textit{Analysis:} The DOI 10.1029/1999JD901133 is valid and corresponds to a real publication, but the title and authors do not match the paper cited in the 'Tool Output'. The actual paper is titled 'Simulations of water vapor...' by Andrew Gettelman et al., which is unrelated to the Southern Annular Mode. The 'Tool Output' incorrectly attributes a paper on SAM to this DOI.
    
    \vspace{2mm}\hrule\vspace{2mm}
    
    \textbf{Qwen3-VL-32B:} \textcolor{red}{[Authentic]} \\
    \textit{Analysis:} The DOI resolves to a real publication... The title and year match exactly. The paper is well-known.
    
    \vspace{2mm}\hrule\vspace{2mm}
    
    \textbf{Gemini-Flash:} \textcolor{red}{[Authentic]} \\
    \textit{Analysis:} The paper by Thompson and Wallace (2000) is a real and influential publication. The DOI correctly points to this title.
\end{promptbox}